%% file: paper.tex
\pdfoutput=1
\documentclass{sysu_preprint}
\DeclareTextFontCommand{\textbf}{\bfseries}
\usepackage{fix-cm}
\usepackage[utf8]{inputenc}
\usepackage{amsmath,amssymb,amsthm,mathtools}
\usepackage{graphicx,booktabs,multirow,tabularx,array,longtable,colortbl,placeins,needspace,float,capt-of}
\usepackage{algorithm,algpseudocode}
\usepackage{tikz}
\usetikzlibrary{arrows.meta,positioning}
\definecolor{baselinebest}{RGB}{225,241,225}
\definecolor{baselinesecond}{gray}{0.93}
\hypersetup{colorlinks=true,linkcolor=blue,citecolor=blue,urlcolor=blue,pdftitle={TeamFlowRL: Online Policy Optimization of Decentralized Flow Policies for Multi-Agent Coordination},pdfauthor={Guowei Zou, Beiwen Zhang, Haitao Wang, Hong Hong, Zhiquan Chen, Hejun Wu}}
\newcommand{\E}{\mathbb{E}}
\newcommand{\KL}{D_{\mathrm{KL}}}
\newcommand{\clip}{\operatorname{clip}}
\newcommand{\N}{\mathcal{N}}

\newtheorem{proposition}{Proposition}
\input{tables/maniskill_frames_meta}

\input{tables/robofactory_frames_meta}

\title{TeamFlowRL: Online Policy Optimization of Decentralized Flow Policies for Multi-Agent Coordination}
\renewcommand{\authorlist}{\authorfont{\sffamily Guowei Zou\textsuperscript{1,*}, Beiwen Zhang\textsuperscript{1,*}, Haitao Wang\textsuperscript{1}, Hong Hong\textsuperscript{1}, Zhiquan Chen\textsuperscript{1}, Hejun Wu\textsuperscript{1,\textdagger}}}
\abstract{Complex cooperative tasks require coordinated multi-agent behavior. Offline generative policies learn expressive cooperative behaviors from trajectories but remain limited by fixed datasets. Online reinforcement learning can improve them through environment interaction. However, fine-tuning flow policies is difficult because action likelihoods are hard to compute. We propose TeamFlowRL, a two-stage framework that pretrains multi-agent flow policies offline and fine-tunes them online.  We formulate team policy optimization over the complete denoising trajectories of decentralized flow policies, avoiding intractable action likelihoods. The resulting objective weights each agent's trajectory ratio with a shared team advantage and clips each ratio separately, aligning local policy updates with the common team return.
Across 51 paired configurations in seven simulation groups, TeamFlowRL improves every offline initialization, with a 47.3\% average gain in normalized score. It also achieves 33.1\% higher average success under exploration noise than Gaussian-policy fine-tuning across 6 bimanual tasks.}
\newcommand{\teaserfigure}{\begin{center}
\centering
\includegraphics[width=.98\linewidth]{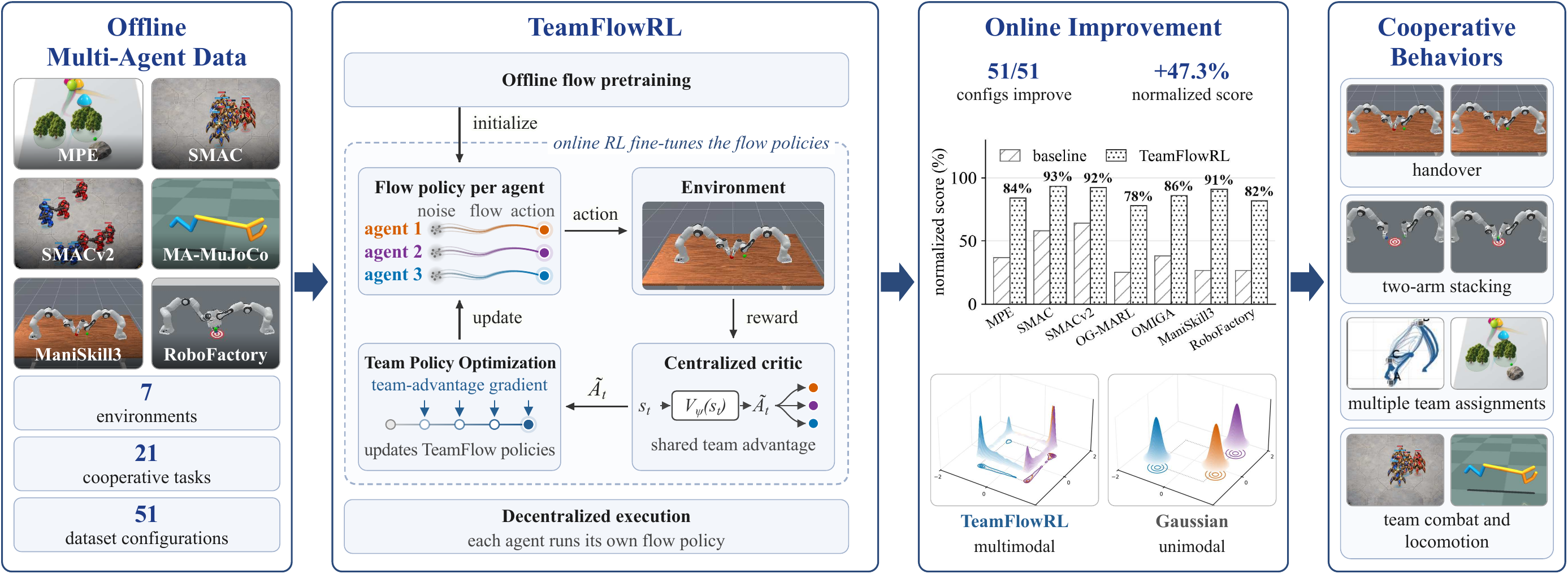}
\captionsetup{hypcap=false}
\captionof{figure}{Overview of TeamFlowRL. Left: 7 environments, 21 cooperative tasks, and 51 dataset configurations. Middle: pretrained flow policies are fine-tuned online with a centralized critic and team policy optimization, then each agent executes its own flow policy from local inputs. Right: all 51 configurations improve (scores normalized per task). Flow actions are multimodal where the Gaussian baseline (BC+MAPPO) is unimodal, and agents learn cooperative behaviors.}
\label{fig:motivation}
\end{center}}
\definecolor{coverpurple}{HTML}{49308C}
\definecolor{covergray}{HTML}{F1F4F8}
\renewcommand{\maketitle}{%
  \thispagestyle{plain}%
  \begin{tcolorbox}[enhanced,colback=covergray,frame hidden,arc=3mm,
    left=6mm,right=6mm,top=4mm,bottom=4mm,boxsep=0pt,before skip=0pt,after skip=5mm]
    {\raggedright\sffamily\bfseries\color{coverpurple}\fontsize{17}{20}\selectfont
      \titlelist\par}
    \vspace{2mm}
    {\raggedright\renewcommand{\authorfont}{\fontsize{10}{12}\selectfont}
      \bfseries\authorlist\par}
    \vspace{2mm}
    {\raggedright\sffamily\bfseries\color{coverpurple}\fontsize{11}{13}\selectfont \textsuperscript{1}Sun Yat-sen University\hfill{\mdseries\color{black!70}\fontsize{8.5}{10}\selectfont \textsuperscript{*}Equal contribution.\quad \textsuperscript{\textdagger}Corresponding author.}\par}
    \vspace{3mm}
    {\color{coverpurple!30}\hrule height 0.3pt}
    \vspace{1mm}
    \teaserfigure
    \vspace{1mm}
    {\color{coverpurple!30}\hrule height 0.3pt}
    \vspace{2mm}
    {\small\hypersetup{urlcolor=coverpurple}
      {\sffamily\bfseries Project Page:} \url{https://teamflowrl.github.io/}\par
      {\sffamily\bfseries Code:} \url{https://github.com/teamflowrl/TeamFlowRL}\par
      {\sffamily\bfseries Models:} \url{https://huggingface.co/teamflowrl/TeamFlowRL}\par}
  \end{tcolorbox}
  {\centering\sffamily\bfseries Abstract\par}
  \vspace{2mm}
  {\setlength{\parindent}{0pt}\abstractlist\par}
  \par\vspace{3mm}
}

\begin{document}
\raggedbottom
\addtocontents{toc}{\protect\setcounter{tocdepth}{-1}}
\maketitle
\input{sections/introduction}

\input{sections/related_work}
\input{sections/method}
\input{sections/experiments}
\input{sections/conclusion}
\clearpage
\section*{LLM Usage Statement}
We used large language model assistants (Claude and Codex) in this work. They were used to polish the English writing and to help implement the method and the plotting code. The authors designed the method and the experiments, verified all code, results, and proofs, and take full responsibility for the content of the paper.
\bibliography{references}
\bibliographystyle{plainnat}
\clearpage
\appendix
\renewcommand{\topfraction}{0.9}\renewcommand{\bottomfraction}{0.8}\renewcommand{\textfraction}{0.07}\renewcommand{\floatpagefraction}{0.85}\setcounter{topnumber}{4}\setcounter{bottomnumber}{3}\setcounter{totalnumber}{6}
\addtocontents{toc}{\protect\setcounter{tocdepth}{2}}
\begingroup
\renewcommand{\contentsname}{Appendix Contents}
\pdfbookmark[0]{Appendix Contents}{appendix-contents}
\tableofcontents
\endgroup
\clearpage
\input{sections/appendix}
\end{document}

%% file: tables/maniskill_frames_meta.tex
\newcommand{\maniskillHasComparison}{1}

%% file: tables/robofactory_frames_meta.tex
\newcommand{\robofactoryFramesReady}{1}

%% file: sections/introduction.tex
\section{Introduction}

Cooperative multi-agent tasks, such as robotic manipulation and cooperative
control, require agents to coordinate their decisions. Multi-agent
reinforcement learning, or MARL, learns such coordination through interaction~\citep{maddpg,qmix,mappo},
while offline generative policies learn expressive action or trajectory
distributions from fixed datasets~\citep{madiff,macflow}.
However, offline policies remain constrained by the quality and coverage of
the available data~\citep{offline_rl_review}.

Online reinforcement learning can further improve coordination through
environment interaction and team rewards. Under centralized training and
decentralized execution, existing policy-gradient methods typically use
categorical distributions for discrete actions or unimodal Gaussians for
continuous actions~\citep{coma,mappo}. Such parameterizations cannot represent
separated action modes for the same observation, whereas offline flow policies
can model richer multimodal behaviors. This motivates fine-tuning expressive
offline flow policies with online team feedback.

Extending single-agent flow policy methods to multi-agent flow policies faces
two severe difficulties. Firstly, a flow policy has no tractable action
probability. PPO-class methods need this probability to compute a ratio of the
new policy and old policy on the same action for policy update and
improvement~\citep{ppo}. As many denoising trajectories of a flow policy reach
the same action, its log-likelihood is intractable and computationally
impossible for multi-agents. Secondly, supposing the ratio can be obtained, the
team update style of existing multi-agent methods, e.g., MAPPO, can no longer be
applied to flow policies directly. In MAPPO, each agent updates its policy
according to the ratio of its new and old policies by a factor of the shared
team advantage~\citep{mappo}. In flow policies, many denoising trajectories of a
flow policy can reach the same action, and the team advantage still evaluates
the final action. Hence, the ratio together with the factor of team advantage is
meaningful only for the final action but not a single trajectory. Even if a
scheme uses the ratio between trajectories instead of between actions, it is
unclear whether this scheme optimizes the team objective or not, since the
relation between a trajectory and a final action is not one to one, but many
to one.

To address these challenges, we propose TeamFlowRL, a framework for online
fine-tuning of offline-trained flow policies of multi-agents. Figure~\ref{fig:motivation} illustrates this framework. We represent each agent's policy through a stochastic denoising trajectory
with tractable likelihoods. We then formulate a team objective that couples
these local trajectories through a shared advantage from a centralized
critic. Each agent's complete trajectory ratio is clipped separately,
rather than clipping the product of all agents' ratios. With exact team
advantages, this surrogate matches the gradient of the joint surrogate at
the collection policy, up to normalization
(Appendix~\ref{app:proofs}). This connects decentralized flow updates to
team return optimization.

We evaluate TeamFlowRL on 51 task--dataset configurations across five
multi-agent benchmark suites and two bimanual manipulation simulators.
Online fine-tuning outperforms the offline initialization in all 51
configurations, with an average gain of 47.3\% in normalized score. We
further compare TeamFlowRL with fine-tuning Gaussian policies. Behavioral
analyses show that the fine-tuned flow policies capture multimodal action
distributions and exhibit diverse cooperative behaviors. On six bimanual
tasks, TeamFlowRL achieves 33.1\% higher average success than BC+MAPPO
under exploration noise.

Our contributions are threefold:
\begin{itemize}
\item To our knowledge, TeamFlowRL is the first framework for online RL
fine-tuning of offline-pretrained multi-agent flow policies.
\item We formulate a team policy objective over decentralized denoising
trajectories, combining a shared team advantage with separate clipping of
each agent's complete trajectory ratio. With exact advantages, the
surrogate matches the gradient of the joint surrogate at the collection
policy, up to normalization.
\item Across 51 configurations, TeamFlowRL consistently improves offline
initializations and outperforms Gaussian policies in multimodal action
expressiveness and online adaptation efficiency.
\end{itemize}

%% file: sections/related_work.tex
\section{Related Work}\label{sec:related-work}

\input{tables/related_work_comparison}

\paragraph{Offline multi-agent policy learning.}
Offline MARL learns cooperative policies from fixed datasets, using behavior
cloning (BC)~\citep{bc} or value-based and conservative objectives such as TD3+BC,
CQL, and BCQ~\citep{td3bc,cql,bcq}, with OMAR and OMIGA providing additional
offline MARL variants~\citep{omar,omiga}. Generative methods
further improve expressiveness: MADiff, DOM2, and DoF use diffusion for
trajectory or decentralized policy learning~\citep{madiff,dom2,dof}, whereas
MAC-Flow learns a joint flow model and distills decentralized actors for
one-step action generation~\citep{macflow}. All remain constrained by the
quality and coverage of the offline data.

\paragraph{Online multi-agent reinforcement learning.}
Online MARL learns coordination through environment interaction using
centralized critics~\citep{maddpg,coma,facmac}, value decomposition
~\citep{vdn,qmix,qplex}, or MAPPO's centralized value estimates with
agent-level PPO updates~\citep{mappo}. These methods typically use categorical
or unimodal Gaussian policies, limiting multimodal action representation.

\paragraph{Offline pretraining and online fine-tuning.}
Online RL can improve pretrained generative policies by optimizing their
denoising processes. DPPO applies PPO to diffusion transitions~\citep{dppo},
while ReinFlow and FPO construct ratios for flow policies using stochastic
denoising trajectories or flow-matching losses~\citep{reinflow,fpo}. Related
work studies fine-tuning generative robot policies~\citep{d2ppo,ogpo} and
vision-language-action (VLA) models~\citep{pirl,pistar06}, but these approaches
remain single-agent and do not address team-level credit assignment for
decentralized multi-agent flow policies.

%% file: tables/related_work_comparison.tex
\begin{table*}[!t]
\centering
\caption{Comparison with existing methods.}
\label{tab:related-work}
\begingroup
\small
\setlength{\tabcolsep}{3pt}
\renewcommand{\arraystretch}{1.08}
\begin{tabularx}{\textwidth}{@{}lXXcc@{}}
\toprule
Method & Policy representation & Policy optimization & Online & Multi-agent \\
\midrule
DPPO~\citep{dppo} & Diffusion & PPO on denoising transitions & \(\checkmark\) & --- \\
ReinFlow~\citep{reinflow} & Flow & PPO on denoising trajectories & \(\checkmark\) & --- \\
FPO~\citep{fpo} & Flow & Flow matching loss ratio & \(\checkmark\) & --- \\
MAPPO~\citep{mappo} & Categorical / Gaussian & PPO on each agent's action & \(\checkmark\) & \(\checkmark\) \\
MADiff~\citep{madiff} & Diffusion & Return conditioning & \(\times\) & \(\checkmark\) \\
DOM2~\citep{dom2} & Diffusion & Q-learning + BC & \(\times\) & \(\checkmark\) \\
MAC-Flow~\citep{macflow} & Actor distilled from flow & Value-guided distillation & \(\times\) & \(\checkmark\) \\
\midrule
\textbf{TeamFlowRL (ours)} & \textbf{Pretrained flow generator} & \textbf{PPO on each agent's denoising trajectory} & \(\checkmark\) & \(\checkmark\) \\
\bottomrule
\end{tabularx}
\endgroup
\end{table*}

%% file: sections/method.tex
\section{Multi-Agent Flow Policy Optimization}
\begin{figure*}[t]
\centering
\includegraphics[width=\textwidth]{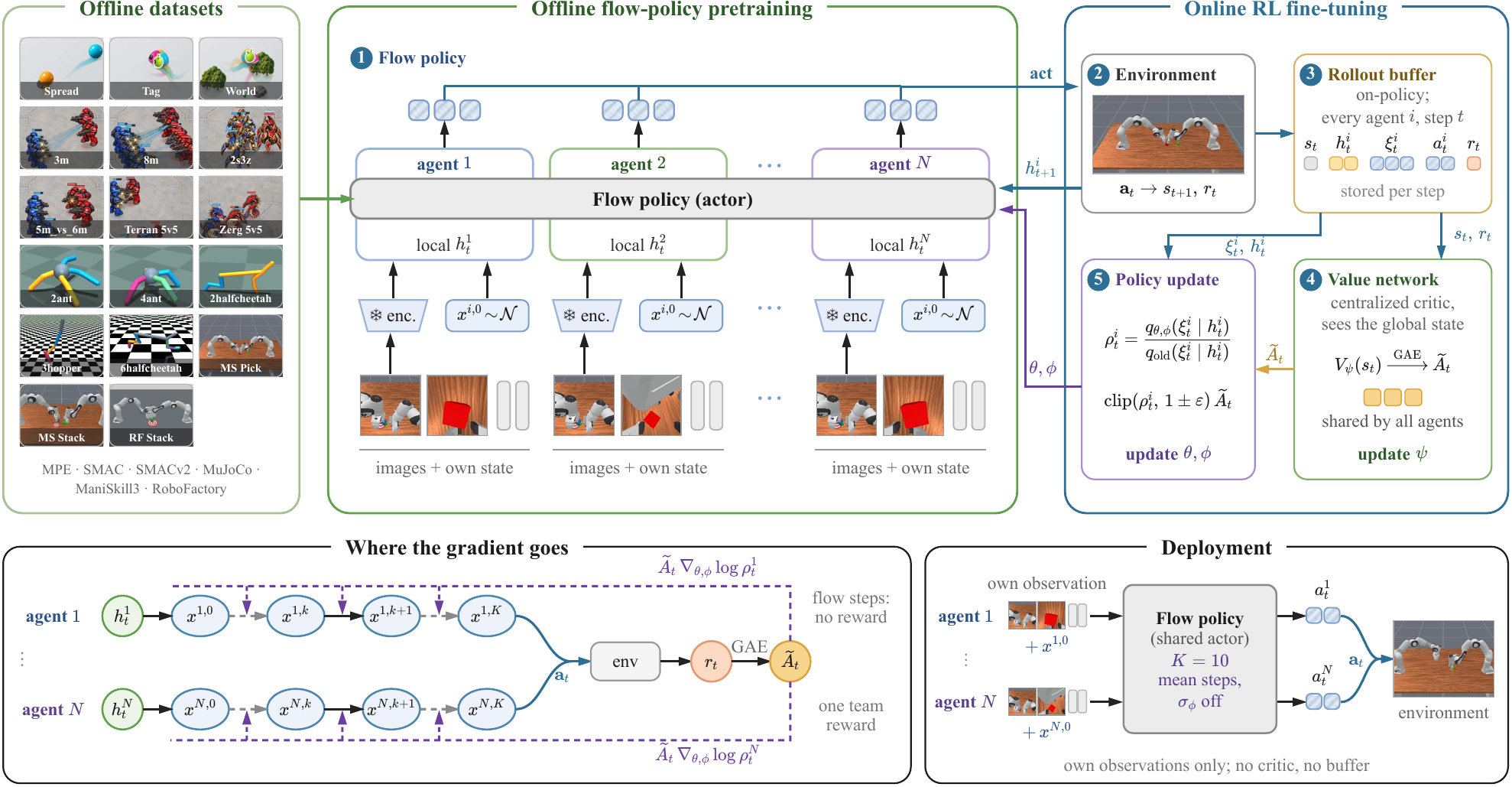}
\caption{TeamFlowRL combines offline flow pretraining (top left and center) with online RL fine-tuning (top right). The bottom left shows how a shared team advantage guides updates along each stored denoising trajectory. The bottom right shows decentralized execution without transition noise.}
\label{fig:framework}
\end{figure*}

TeamFlowRL learns decentralized policies for cooperative tasks in two stages.
Figure~\ref{fig:framework} illustrates the complete procedure. It first
initializes a shared flow policy from offline input--action pairs. For online
fine-tuning, we turn the flow generator into a stochastic policy over complete
denoising trajectories, making the likelihood of each agent's sampled
trajectory tractable. A centralized critic then estimates a shared team
advantage, which multiplies the agent-wise trajectory likelihood ratios in a PPO
objective and jointly updates the shared velocity and noise networks. At
deployment, each agent independently generates its action from local
information using the fine-tuned velocity network.

\subsection{\texorpdfstring{Cooperative problem formulation}{Cooperative problem formulation}}
We model a cooperative task as a decentralized partially observable
Markov decision process, abbreviated Dec-POMDP,
$\langle\mathcal I,\mathcal S,\{\mathcal A^i\}_{i=1}^N,P,R,
\{\mathcal O^i\}_{i=1}^N,O,\gamma\rangle$,
where $\mathcal I=\{1,\ldots,N\}$ is the agent set,
$\mathcal S$ is the global state space,
$\mathcal A^i$ and $\mathcal O^i$ are the local action and observation spaces,
$P$ is the state transition distribution,
$R$ is the common reward function,
$O$ is the observation distribution,
and $\gamma\in(0,1)$ is the discount factor.
At environment step $t$, agent $i$ selects action $a_t^i$
using its local information $h_t^i$, and all agents receive
the shared reward $r_t$.
The local information $h_t^i$ contains only features available to agent $i$.
The goal is to learn a joint action policy $\boldsymbol\pi$ that maximizes the expected discounted team return, $\E_{\boldsymbol\pi,P}[\sum_{t\ge0}\gamma^t r_t]$.

\subsection{Offline flow initialization}\label{sec:offline-init}
We initialize a shared velocity network $v_\theta$ from offline
input--action pairs $(h^i,a^i)\in\mathcal D$. Here and within a single
denoising trajectory, we omit environment time $t$.

\begingroup
Conditional flow matching~\citep{flow} trains $v_\theta$ to predict the velocity along a straight interpolation from Gaussian noise to the data action:
\begin{align}
x_\tau^i&=(1-\tau)z^i+\tau a^i,\qquad z^i\sim\N(0,I),\label{eq:interp}\\
\mathcal L_{\mathrm{FM}}(\theta)&=\E_{\mathcal D,z,\tau,i}\!\left[\|v_\theta(h^i,x_\tau^i,\tau)-(a^i-z^i)\|_2^2\right],\label{eq:fm}
\end{align}
where $\tau\sim\mathcal U[0,1]$ denotes flow time. For discrete action spaces, $a^i$ in Equations~\ref{eq:interp} and~\ref{eq:fm} is the one-hot action vector.\endgroup{}
The initialized policy generates actions by integrating $v_\theta$ from
a Gaussian latent with $K$ Euler steps.

\subsection{A stochastic policy over denoising trajectories}\label{sec:path-policy}
Conditioned on the local input and initial latent, the Euler flow
trajectory is deterministic and does not provide tractable transition
likelihoods for PPO. Following ReinFlow, we add
Gaussian noise to each denoising transition, yielding a tractable
probability over the complete trajectory.

\begingroup
At each environment step, agent $i$ samples $x^{i,0}\sim\N(0,I)$ and applies the following stochastic update for $k=0,\ldots,K-1$:
\begin{equation}
\begin{aligned}
x^{i,k+1}&=\mu_\theta(h^i,x^{i,k},k)+\sigma_\phi(h^i,x^{i,k},k)\odot\epsilon^{i,k},\\
\mu_\theta(h^i,x^{i,k},k)&=x^{i,k}+K^{-1}v_\theta(h^i,x^{i,k},k/K),
\end{aligned}
\label{eq:transition}
\end{equation}
with independent $\epsilon^{i,k}\sim\N(0,I)$ and coordinatewise product
$\odot$. The mean retains the pretrained Euler update, while a shared
noise network learns bounded, positive standard deviations $\sigma_\phi$.

The complete denoising trajectory $\xi^i=(x^{i,0},\ldots,x^{i,K})$ therefore has the joint probability
\begin{equation}
\begin{aligned}q_{\theta,\phi}(\xi^i\mid h^i)={}&p_0(x^{i,0})\\&\cdot\prod_{k=0}^{K-1}\N\!\bigl(x^{i,k+1}\mid\mu_\theta^{i,k},\operatorname{diag}((\sigma_\phi^{i,k})^2)\bigr),\end{aligned}
\label{eq:path}
\end{equation}
where $p_0$ is the standard Gaussian probability and $\mu_\theta^{i,k},\sigma_\phi^{i,k}$ abbreviate the two functions in Equation~\ref{eq:transition}.\endgroup{}
The environment receives only $a^i=g(x^{i,K})$, where $g$ clips continuous
actions or selects the largest legal coordinate for discrete actions.
This trajectory likelihood provides a tractable alternative to the action marginal for policy optimization.

Given local inputs, agents sample their initial latents and transition
noise independently. Despite parameter sharing, the joint trajectory sampling
policy therefore factorizes. We denote this joint trajectory policy by
$M_{\theta,\phi}$ (and reserve $q_{\theta,\phi}$ for one agent), giving
the team objective
\begin{equation}
\begin{aligned}
M_{\theta,\phi}(\boldsymbol\xi_t\mid\mathbf h_t)&=\prod_{i=1}^{N}q_{\theta,\phi}(\xi_t^i\mid h_t^i),\\
J(\theta,\phi)&=\E_{M_{\theta,\phi}}\Bigl[\sum_{t\ge 0}\gamma^t r_t\Bigr].
\end{aligned}
\label{eq:joint}
\end{equation}
Here, $\boldsymbol\xi_t=(\xi_t^1,\ldots,\xi_t^N)$ denotes the joint denoising trajectory, and $\mathbf h_t=(h_t^1,\ldots,h_t^N)$ denotes the tuple of actor inputs.

\subsection{Team advantages and policy updates}\label{sec:team-update}
The remaining step is to connect individual trajectory likelihoods to team
returns. The factorization in Equation~\ref{eq:joint} gives
\begin{equation}
{
\nabla_{\theta,\phi}\log M_{\theta,\phi}(\boldsymbol\xi_t\mid\mathbf h_t)
=
\sum_{i=1}^{N}
\nabla_{\theta,\phi}\log q_{\theta,\phi}(\xi_t^i\mid h_t^i).
}
\label{eq:joint-path-score}
\end{equation}
The team policy gradient multiplies this sum by a common team advantage. We use centralized advantage estimation and PPO clipping as in MAPPO, with a likelihood ratio for each denoising trajectory.

The centralized critic $V_\psi(s_t)$ estimates team returns from the global
state. GAE~\citep{gae} gives one advantage for each environment transition.
We normalize it over the rollout and assign the same team advantage
$\widetilde A_t$ to every active agent. The critic target is
$\widehat R_t=\widehat A_t+V_{\psi_{\mathrm{old}}}(s_t)$, and we fit
$V_\psi$ to this fixed target with the standard squared error objective.

Each agent's update uses its complete stored denoising trajectory:
\begin{equation}
\rho_t^i=\frac{q_{\theta,\phi}(\xi_t^i\mid h_t^i)}{q_{\mathrm{old}}(\xi_t^i\mid h_t^i)},\qquad Z=\max\Bigl(1,\sum_{t,i}m_t^i\Bigr),
\label{eq:ratio}
\end{equation}
where $q_{\mathrm{old}}$ is the rollout policy and $m_t^i\in\{0,1\}$ marks
agents with an action choice. The log ratio sums transition log probabilities over all denoising steps and action coordinates. The fixed initial probability $p_0$ cancels.

Clipping the product of all agents' ratios makes clipping depend on their
combined change. We instead clip each agent's complete trajectory ratio
separately, while retaining the shared team advantage, and average over
active samples in the rollout:
\begin{equation}
L_{\mathrm{clip}}=\frac{1}{Z}\sum_{t,i}m_t^i\min\Bigl(\rho_t^i\widetilde A_t,\ \clip(\rho_t^i,1-\varepsilon,1+\varepsilon)\widetilde A_t\Bigr),
\label{eq:ppo}
\end{equation}
where $\varepsilon$ is the clipping threshold. With exact team advantages,
the two surrogates have the same first-order direction at the collection
policy, up to the common scale $1/Z$. Appendix~\ref{app:method-details}
gives the noise parameterization, the formal connection to the action
marginal, and the theoretical analysis of this update.

We minimize $-L_{\mathrm{clip}}$ while holding the stored denoising
trajectories fixed. Gradients update $\theta$ and $\phi$ through the
conditional means and standard deviations.
Appendix~\ref{app:part-other} gives the GAE recursion, boundary handling,
and implementation details.

\input{tables/main_results_compact}

\paragraph{Deployment.}

Each agent samples a fresh Gaussian latent and applies the $K$ mean updates
in Equation~\ref{eq:transition} using only $h_t^i$. Training uses transition
noise, while deployment removes this noise and requires neither the noise
network nor the centralized critic. The initial latent keeps the executed
policy stochastic.

%% file: tables/main_results_compact.tex
\begin{table*}[!h]
\centering
\caption{Results on MPE and OG-MARL MuJoCo (continuous control) and SMAC (discrete control), with methods as rows and task--dataset pairs as columns. Local reproductions are marked $^{\mathrm r}$ and are not directly comparable to the published MAC-Flow configurations. OMIGA MuJoCo and SMACv2 results are in Table~\ref{tab:omiga-smacv2}. Shading marks the best and second-best value in each column. Sources and notation are in Appendix~\ref{app:experimental-protocol}.}
\label{tab:main-results}
\label{tab:main-results-compact}
\label{tab:continuous-main}
\label{tab:discrete-main}
\label{tab:published_continuous}
\label{tab:published_discrete}
\label{tab:mujoco_combined}
\label{tab:mujoco_ogmarl}
\begingroup\scriptsize
\renewcommand{\tiny}{\fontsize{5}{6}\selectfont}
\setlength{\tabcolsep}{1.5pt}
\renewcommand{\arraystretch}{1.12}
\resizebox{\linewidth}{!}{%
\begin{tabular}{@{}lcccccccccc@{}}
\toprule
\multirow{2}{*}{\textbf{MPE}} & \multicolumn{4}{c}{\textbf{Spread}} & \multicolumn{3}{c}{\textbf{Tag}} & \multicolumn{3}{c}{\textbf{World}} \\
\cmidrule(lr){2-5}\cmidrule(lr){6-8}\cmidrule(lr){9-11}
 & Expert & Medium & M-R$^{b}$ & Random & Expert & Medium & Random & Expert & Medium & Random \\
\midrule
MADiff-D$^{a}$~\citep{madiff}~(\citeyear{madiff}) & $95.0$\,{\tiny$\pm 5.3$} & $64.9$\,{\tiny$\pm 7.7$} & $30.3$\,{\tiny$\pm 2.5$} & $6.9$\,{\tiny$\pm 3.1$} & $120.9$\,{\tiny$\pm 14.6$}$^{\mathrm{m}}$ & $77.2$\,{\tiny$\pm 10.4$}$^{\mathrm{m}}$ & $3.2$\,{\tiny$\pm 4.0$}$^{\mathrm{m}}$ & $122.6$\,{\tiny$\pm 14.4$}$^{\mathrm{m}}$ & \cellcolor{baselinesecond}\underline{${123.5}$\,{\tiny$\pm 4.5$}$^{\mathrm{m}}$} & $2.0$\,{\tiny$\pm 3.0$}$^{\mathrm{m}}$ \\
DOM2~\citep{dom2}~(\citeyear{dom2}) & \cellcolor{baselinebest}{\boldmath ${131.32}$\,{\tiny$\pm 4.82$}} & $55.77$\,{\tiny$\pm 7.06$} & \cellcolor{baselinesecond}\underline{${46.02}$\,{\tiny$\pm 10.81$}} & \cellcolor{baselinebest}{\boldmath ${49.86}$\,{\tiny$\pm 7.28$}} & \cellcolor{baselinesecond}\underline{${138.75}$\,{\tiny$\pm 12.02$}} & \cellcolor{baselinesecond}\underline{${84.29}$\,{\tiny$\pm 25.36$}} & \cellcolor{baselinesecond}\underline{${112.18}$\,{\tiny$\pm 30.21$}} & \cellcolor{baselinesecond}\underline{${123.17}$\,{\tiny$\pm 19.81$}} & $105.79$\,{\tiny$\pm 27.11$} & $54.23$\,{\tiny$\pm 16.57$} \\
MAC-Flow~\citep{macflow}~(\citeyear{macflow}) & $101.7$\,{\tiny$\pm 10.9$} & \cellcolor{baselinebest}{\boldmath ${80.1}$\,{\tiny$\pm 20.6$}} & \cellcolor{baselinebest}{\boldmath ${50.4}$\,{\tiny$\pm 33.2$}} & \cellcolor{baselinesecond}\underline{$31.1$\,{\tiny$\pm 6.8$}} & $119.97^{\mathrm{r}}$\,{\tiny$\pm 2.26$} & $80.47^{\mathrm{r}}$\,{\tiny$\pm 1.77$} & $72.68^{\mathrm{r}}$\,{\tiny$\pm 3.20$} & $122.57^{\mathrm{r}}$\,{\tiny$\pm 2.91$} & $89.82^{\mathrm{r}}$\,{\tiny$\pm 2.25$} & \cellcolor{baselinesecond}\underline{${94.43}^{\mathrm{r}}$\,{\tiny$\pm 3.72$}} \\
\midrule
TeamFlowRL (w/o online RL) & $72.87$\,{\tiny$\pm 4.85$} & $27.13$\,{\tiny$\pm 0.93$} & $7.89$\,{\tiny$\pm 4.79$} & $-1.79$\,{\tiny$\pm 3.45$} & $96.60$\,{\tiny$\pm 7.31$} & $58.62$\,{\tiny$\pm 6.47$} & $-0.40$\,{\tiny$\pm 0.99$} & $109.23$\,{\tiny$\pm 13.01$} & $61.69$\,{\tiny$\pm 6.60$} & $0.18$\,{\tiny$\pm 1.60$} \\
\textbf{TeamFlowRL} & \cellcolor{baselinesecond}{\underline{${110.75}^{\dagger}$\,{\tiny$\pm 3.76$}}} & \cellcolor{baselinesecond}{\underline{${71.13}^{\dagger}$\,{\tiny$\pm 8.31$}}} & ${31.32}^{\dagger}$\,{\tiny$\pm 7.54$} & $24.93^{\dagger}$\,{\tiny$\pm 2.31$} & \cellcolor{baselinebest}{\boldmath ${150.13}^{\dagger}$\,{\tiny$\pm 5.83$}} & \cellcolor{baselinebest}{\boldmath ${145.13}^{\dagger}$\,{\tiny$\pm 7.88$}} & \cellcolor{baselinebest}{\boldmath ${156.74}^{\dagger}$\,{\tiny$\pm 4.45$}} & \cellcolor{baselinebest}{\boldmath ${176.11}^{\dagger}$\,{\tiny$\pm 16.62$}} & \cellcolor{baselinebest}{\boldmath ${155.60}^{\dagger}$\,{\tiny$\pm 4.71$}} & \cellcolor{baselinebest}{\boldmath ${169.15}^{\dagger}$\,{\tiny$\pm 16.77$}} \\
\bottomrule
\end{tabular}}
\par\vspace{3pt}
\resizebox{\linewidth}{!}{%
\begin{tabular}{@{}lccccccccc@{}}
\toprule
\multirow{2}{*}{\textbf{OG-MARL}} & \multicolumn{3}{c}{\textbf{2ant}} & \multicolumn{3}{c}{\textbf{2halfcheetah}} & \multicolumn{3}{c}{\textbf{4ant}} \\
\cmidrule(lr){2-4}\cmidrule(lr){5-7}\cmidrule(lr){8-10}
 & Good & Medium & Poor & Good & Medium & Poor & Good & Medium & Poor \\
\midrule
MADiff-D$^{a}$~\citep{madiff}~(\citeyear{madiff}) & \cellcolor{baselinesecond}\underline{$2946$\,{\tiny$\pm 77$}} & $1211$\,{\tiny$\pm 69$} & $946$\,{\tiny$\pm 66$} & $8246$\,{\tiny$\pm 342$} & $2207$\,{\tiny$\pm 23$} & $759$\,{\tiny$\pm 18$} & \cellcolor{baselinesecond}\underline{$3080$\,{\tiny$\pm 38$}} & $1649$\,{\tiny$\pm 100$} & $1295$\,{\tiny$\pm 57$} \\
OMSD~\citep{omsd}~(\citeyear{omsd}) & $2714$\,{\tiny$\pm 555$} & \cellcolor{baselinesecond}\underline{$1372$\,{\tiny$\pm 107$}} & \cellcolor{baselinesecond}\underline{$1213$\,{\tiny$\pm 212$}} & \cellcolor{baselinesecond}\underline{$8619$\,{\tiny$\pm 418$}} & \cellcolor{baselinesecond}\underline{$2660$\,{\tiny$\pm 125$}} & $866$\,{\tiny$\pm 78$} & $2844$\,{\tiny$\pm 152$} & \cellcolor{baselinesecond}\underline{$1942$\,{\tiny$\pm 293$}} & \cellcolor{baselinesecond}\underline{$1477$\,{\tiny$\pm 192$}} \\
MAC-Flow$^{\mathrm{r}}$~\citep{macflow}~(\citeyear{macflow}) & $1628.46^{\mathrm{r}}$\,{\tiny$\pm 1.72$} & $1077.94^{\mathrm{r}}$\,{\tiny$\pm 42.96$} & $912.21^{\mathrm{r}}$\,{\tiny$\pm 23.51$} & ${8569.00}^{\mathrm{r}}$\,{\tiny$\pm 42.59$} & $2477.22^{\mathrm{r}}$\,{\tiny$\pm 4.11$} & \cellcolor{baselinesecond}\underline{${937.09}^{\mathrm{r}}$\,{\tiny$\pm 8.23$}} & $1750.85^{\mathrm{r}}$\,{\tiny$\pm 29.29$} & $1254.43^{\mathrm{r}}$\,{\tiny$\pm 35.45$} & $1010.19^{\mathrm{r}}$\,{\tiny$\pm 39.19$} \\
\midrule
TeamFlowRL (w/o online RL) & $527.15$\,{\tiny$\pm 134.37$} & $956.05$\,{\tiny$\pm 50.61$} & $535.21$\,{\tiny$\pm 6.77$} & $6204.54$\,{\tiny$\pm 43.82$} & $1616.70$\,{\tiny$\pm 33.86$} & $544.34$\,{\tiny$\pm 98.30$} & $1554.84$\,{\tiny$\pm 41.54$} & $1263.13$\,{\tiny$\pm 42.48$} & $515.18$\,{\tiny$\pm 73.05$} \\
\textbf{TeamFlowRL} & \cellcolor{baselinebest}{\boldmath ${3453.45}$\,{\tiny$\pm 224.71$}} & \cellcolor{baselinebest}{\boldmath ${3000.60}$\,{\tiny$\pm 53.38$}} & \cellcolor{baselinebest}{\boldmath ${3283.65}$\,{\tiny$\pm 95.53$}} & \cellcolor{baselinebest}{\boldmath ${10559.68}$\,{\tiny$\pm 167.82$}} & \cellcolor{baselinebest}{\boldmath ${4568.23}$\,{\tiny$\pm 93.63$}} & \cellcolor{baselinebest}{\boldmath ${1414.51}$\,{\tiny$\pm 13.46$}} & \cellcolor{baselinebest}{\boldmath ${3769.81}$\,{\tiny$\pm 36.64$}} & \cellcolor{baselinebest}{\boldmath ${3124.54}$\,{\tiny$\pm 7.83$}} & \cellcolor{baselinebest}{\boldmath ${2957.69}$\,{\tiny$\pm 74.82$}} \\
\bottomrule
\end{tabular}}
\par\vspace{3pt}
\resizebox{\linewidth}{!}{%
\begin{tabular}{@{}lcccccccccccc@{}}
\toprule
\multirow{2}{*}{\textbf{SMACv1}} & \multicolumn{3}{c}{\textbf{3m}} & \multicolumn{3}{c}{\textbf{8m}} & \multicolumn{3}{c}{\textbf{2s3z}} & \multicolumn{3}{c}{\textbf{5m6m}} \\
\cmidrule(lr){2-4}\cmidrule(lr){5-7}\cmidrule(lr){8-10}\cmidrule(lr){11-13}
 & Good & Medium & Poor & Good & Medium & Poor & Good & Medium & Poor & Good & Medium & Poor \\
\midrule
Diffusion BC~\citep{diffusion_policy}~(\citeyear{diffusion_policy}) & $19.5$\,{\tiny$\pm 0.5$} & $13.3$\,{\tiny$\pm 0.7$} & $4.2$\,{\tiny$\pm 0.2$} & $19.4$\,{\tiny$\pm 0.5$} & $18.6$\,{\tiny$\pm 0.6$} & $4.8$\,{\tiny$\pm 0.2$} & $18.0$\,{\tiny$\pm 1.0$} & $13.4$\,{\tiny$\pm 1.4$} & $6.2$\,{\tiny$\pm 1.2$} & $16.8$\,{\tiny$\pm 2.3$} & $12.5$\,{\tiny$\pm 2.1$} & $8.0$\,{\tiny$\pm 1.0$} \\
Flow BC~\citep{flow}~(\citeyear{flow}) & \cellcolor{baselinebest}{\boldmath ${20.0}$\,{\tiny$\pm 0.0$}} & $14.7$\,{\tiny$\pm 1.5$} & $4.5$\,{\tiny$\pm 0.1$} & $19.5$\,{\tiny$\pm 0.2$} & $18.2$\,{\tiny$\pm 0.8$} & $4.9$\,{\tiny$\pm 0.1$} & \cellcolor{baselinesecond}\underline{${19.5}$\,{\tiny$\pm 0.1$}} & $15.1$\,{\tiny$\pm 2.0$} & $6.9$\,{\tiny$\pm 0.8$} & $14.7$\,{\tiny$\pm 2.1$} & $12.8$\,{\tiny$\pm 0.8$} & $7.7$\,{\tiny$\pm 0.8$} \\
MADiff-D$^{a}$~\citep{madiff}~(\citeyear{madiff}) & $19.3$\,{\tiny$\pm 0.5$} & $16.4$\,{\tiny$\pm 2.6$} & $10.3$\,{\tiny$\pm 6.1$} & $18.9$\,{\tiny$\pm 1.1$} & $16.8$\,{\tiny$\pm 1.6$} & $9.8$\,{\tiny$\pm 0.9$} & $15.9$\,{\tiny$\pm 1.2$} & $15.6$\,{\tiny$\pm 0.3$} & $8.5$\,{\tiny$\pm 1.3$} & $16.5$\,{\tiny$\pm 2.8$} & $15.2$\,{\tiny$\pm 2.6$} & $8.9$\,{\tiny$\pm 1.3$} \\
DoF~\citep{dof}~(\citeyear{dof}) & \cellcolor{baselinesecond}\underline{${19.8}$\,{\tiny$\pm 0.2$}} & \cellcolor{baselinesecond}\underline{${18.6}$\,{\tiny$\pm 1.2$}} & \cellcolor{baselinesecond}\underline{${10.9}$\,{\tiny$\pm 1.1$}} & $19.6$\,{\tiny$\pm 0.3$} & $18.6$\,{\tiny$\pm 0.8$} & \cellcolor{baselinesecond}\underline{${12.0}$\,{\tiny$\pm 1.2$}} & $18.5$\,{\tiny$\pm 0.8$} & \cellcolor{baselinesecond}\underline{${18.1}$\,{\tiny$\pm 0.9$}} & \cellcolor{baselinesecond}\underline{${10.0}$\,{\tiny$\pm 1.1$}} & \cellcolor{baselinesecond}\underline{${17.7}$\,{\tiny$\pm 1.1$}} & \cellcolor{baselinesecond}\underline{${16.2}$\,{\tiny$\pm 0.9$}} & \cellcolor{baselinesecond}\underline{${10.8}$\,{\tiny$\pm 0.3$}} \\
MAC-Flow~\citep{macflow}~(\citeyear{macflow}) & \cellcolor{baselinesecond}\underline{${19.8}$\,{\tiny$\pm 0.2$}} & $18.0$\,{\tiny$\pm 3.2$} & $10.6$\,{\tiny$\pm 2.2$} & \cellcolor{baselinesecond}\underline{${19.7}$\,{\tiny$\pm 0.3$}} & \cellcolor{baselinebest}{\boldmath ${19.4}$\,{\tiny$\pm 0.6$}} & $11.5$\,{\tiny$\pm 0.8$} & \cellcolor{baselinesecond}\underline{${19.5}$\,{\tiny$\pm 0.5$}} & $17.6$\,{\tiny$\pm 0.6$} & $8.5$\,{\tiny$\pm 0.6$} & \cellcolor{baselinebest}{\boldmath ${18.6}$\,{\tiny$\pm 3.5$}} & $15.6$\,{\tiny$\pm 1.3$} & $9.8$\,{\tiny$\pm 2.1$} \\
\midrule
TeamFlowRL (w/o online RL) & $15.85$\,{\tiny$\pm 1.28$} & $10.12$\,{\tiny$\pm 1.11$} & $4.26$\,{\tiny$\pm 0.13$} & $17.87$\,{\tiny$\pm 0.44$} & $11.19$\,{\tiny$\pm 1.26$} & $5.20$\,{\tiny$\pm 0.11$} & $18.28$\,{\tiny$\pm 0.54$} & $12.11$\,{\tiny$\pm 0.54$} & $7.12$\,{\tiny$\pm 0.35$} & $15.50$\,{\tiny$\pm 0.38$} & $11.57$\,{\tiny$\pm 0.74$} & $8.01$\,{\tiny$\pm 0.47$} \\
\textbf{TeamFlowRL} & \cellcolor{baselinesecond}\underline{${19.80}$\,{\tiny$\pm 0.35$}} & \cellcolor{baselinebest}{\boldmath ${19.81}$\,{\tiny$\pm 0.34$}} & \cellcolor{baselinebest}{\boldmath ${18.63}$\,{\tiny$\pm 0.89$}} & \cellcolor{baselinebest}{\boldmath ${20.00}$\,{\tiny$\pm 0.00$}} & \cellcolor{baselinesecond}\underline{${18.89}$\,{\tiny$\pm 0.64$}} & \cellcolor{baselinebest}{\boldmath ${18.40}$\,{\tiny$\pm 0.34$}} & \cellcolor{baselinebest}{\boldmath ${19.84}$\,{\tiny$\pm 0.15$}} & \cellcolor{baselinebest}{\boldmath ${20.00}$\,{\tiny$\pm 0.00$}} & \cellcolor{baselinebest}{\boldmath ${16.46}$\,{\tiny$\pm 0.67$}} & $17.43$\,{\tiny$\pm 0.77$} & \cellcolor{baselinebest}{\boldmath ${19.02}$\,{\tiny$\pm 0.94$}} & \cellcolor{baselinebest}{\boldmath ${12.85}$\,{\tiny$\pm 0.45$}} \\
\bottomrule
\end{tabular}}
\endgroup
\par\vspace{1pt}
{\scriptsize\itshape\raggedright $^{a}$\,MADiff-D is the decentralized MADiff variant, which matches decentralized execution. $^{b}$\,Medium-Replay (M-R) labels a dataset quality. The listed baselines differ across the three blocks because each method supports or reports only some of the continuous and discrete benchmarks.\par}
\end{table*}

%% file: sections/experiments.tex
\section{Experiments}

We evaluate TeamFlowRL on standard multi-agent benchmarks and cooperative bimanual manipulation. The experiments measure improvement over the offline initialization across tasks and data qualities, examine the action distributions and cooperative behaviors of the fine-tuned policy, compare TeamFlowRL with Gaussian policies, and test the main implementation choices for the trajectory ratio.

\begin{figure*}[!t]\centering
\includegraphics[width=0.95\textwidth]{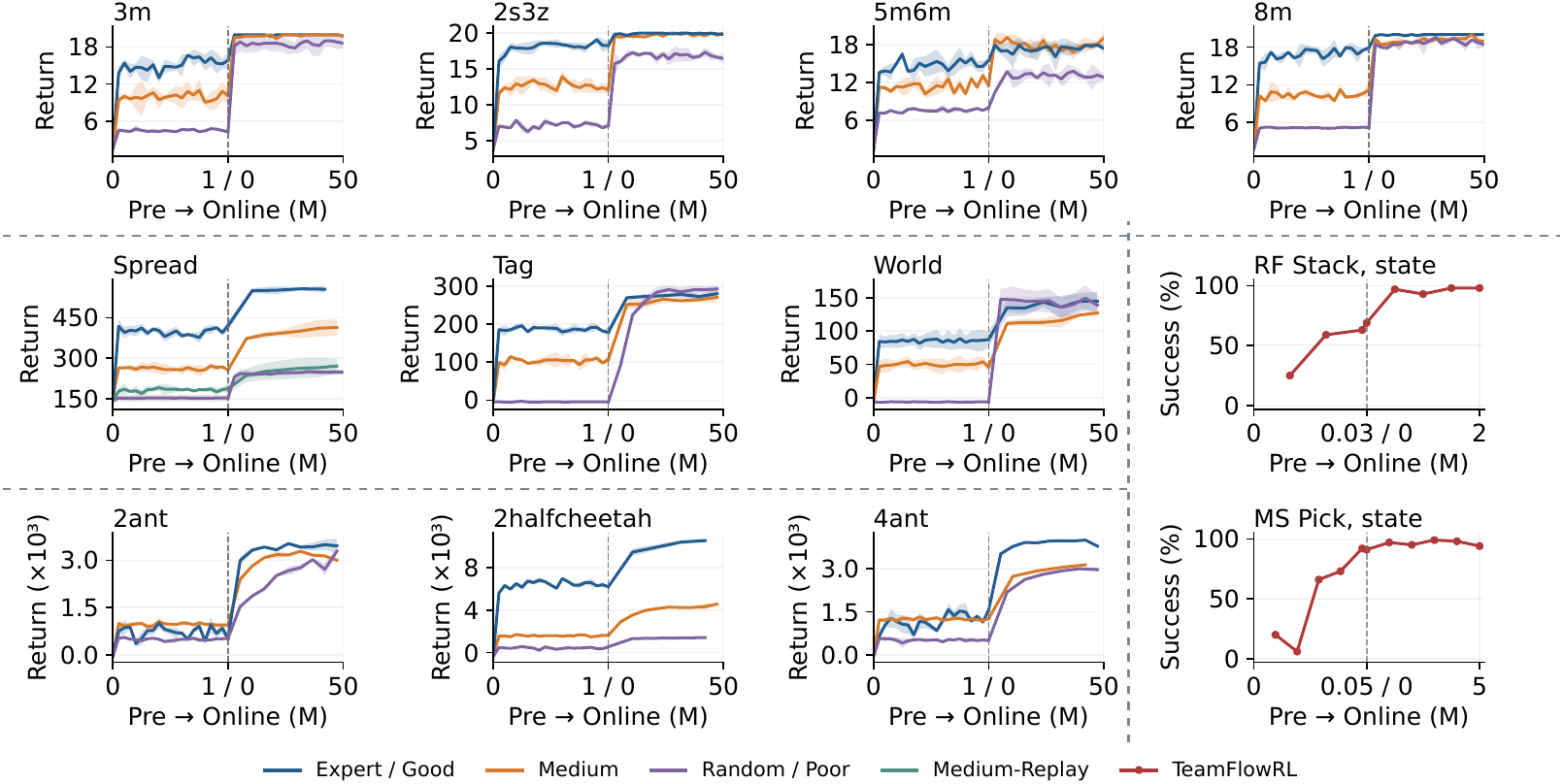}
\caption{Learning curves on SMAC, MPE, OG-MARL MuJoCo, and the robot tasks. The horizontal axis shows pretraining updates followed by online environment steps in millions, and the tick 1 / 0 marks the start of online training. Colors identify offline data quality. Robot panels show the success of TeamFlowRL under the deployment rule. Additional curves and reporting details are provided in Appendix~\ref{app:part-experiments}.\label{fig:robot-learning}}\label{fig:learning}
\end{figure*}
\subsection{Experimental Setup}
\textbf{Benchmarks.} We evaluate five multi-agent benchmark groups: cooperative navigation and pursuit in the Multi-Agent Particle Environment (MPE), OG-MARL MuJoCo~\citep{ogmarl}, OMIGA MuJoCo~\citep{omiga}, SMAC~\citep{smac}, and SMACv2~\citep{smacv2}. They contain 45 combinations of task and dataset. We also evaluate simulated bimanual manipulation using TwoRobotPickCube-v1 and TwoRobotStackCube-v1 from ManiSkill3~\citep{maniskill3} and TwoRobotsStackCube from RoboFactory~\citep{robofactory}, each with two Franka Panda arms, and treat each arm as one agent. We use both state and camera observations and also include a hands-away variant of the two stacking tasks, in which both arms must release the blocks and move away at the end. Complete task settings are provided in Appendix~\ref{app:part-experiments}, with the bimanual configurations and pretraining data in Appendix~\ref{app:robot-setup}.

\textbf{Baselines.} We compare TeamFlowRL with representative
methods based on diffusion and flow models, including DOM2,
MADiff, DoF, Diffusion BC , Flow BC, and MAC-Flow.
The baselines vary across MPE, OG-MARL, and SMAC, as
not all methods support every benchmark.
For bimanual manipulation, we additionally evaluate MAPPO and BC+MAPPO under
matched task definitions, observations, rewards, and online interaction
budgets. Published RoboFactory results use a different visuomotor
imitation-learning protocol and are therefore not directly comparable.

\textbf{Evaluation protocol.} All benchmark results use a common online budget of 50 million environment steps. The robot methods use matched online interaction budgets within each task. The main robot comparison uses stochastic execution, which keeps the exploration noise used in training. All robot results are averaged over three independent training seeds, and the shaded regions denote 95\% confidence intervals over evaluation episodes.
Complete reporting conventions, evaluation details, and results under each method's deployment rule are provided in Appendices~\ref{app:part-experiments} and~\ref{app:part-robot}.

\subsection{Main Results}\label{sec:offline-improvement}

\textbf{Improvement over offline initialization.} Online fine-tuning improves the offline flow policy in all 51 configurations. In all 45 benchmark configurations, the final mean return exceeds that of the offline initialization by 43.2\% of the task's best online return on average. Table~\ref{tab:main-results} reports the main benchmark results. In all six robot tasks, the final success under stochastic execution exceeds that of the pretrained initialization by 61.2\% on average, as shown in Table~\ref{tab:robot-results}. We normalize benchmark returns by the best online return in Figure~\ref{fig:motivation} and express robot success under exploration noise as percentages. The average gain in normalized score across the seven groups and all 51 configurations is 47.3\%.

\begin{figure}[!t]
\centering
\includegraphics[width=0.55\linewidth]{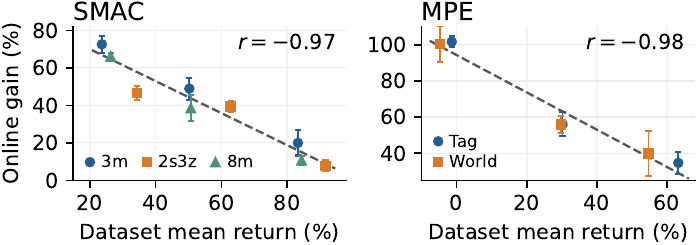}
\caption{Online gain versus offline dataset mean return, both as a percentage of the task's best online return. Dashed lines are linear fits, and error bars show one standard deviation.}\label{fig:quality-gain}
\end{figure}

\textbf{Effect of offline data quality.} Figure~\ref{fig:quality-gain} shows
that gains occur at every data quality and shrink linearly as the offline data
improve. Over the 15 SMAC and MPE configurations, each 1\% increase in dataset
mean return lowers the online gain by 0.93\%, with Pearson correlation
$r=-0.98$. In 3m, 2s3z, 8m, Tag, and World, the policy fine-tuned from the
Good or Expert data exceeds the mean return of that dataset. In 3m, 8m, Tag,
and World, the policy fine-tuned from the Poor or Random data ends within 10\%
of the best final return. Additional learning curves and robot episodes are provided in Appendices~\ref{app:part-experiments} and~\ref{app:part-robot}.

\begin{figure*}[!t]\centering
\includegraphics[width=0.88\textwidth]{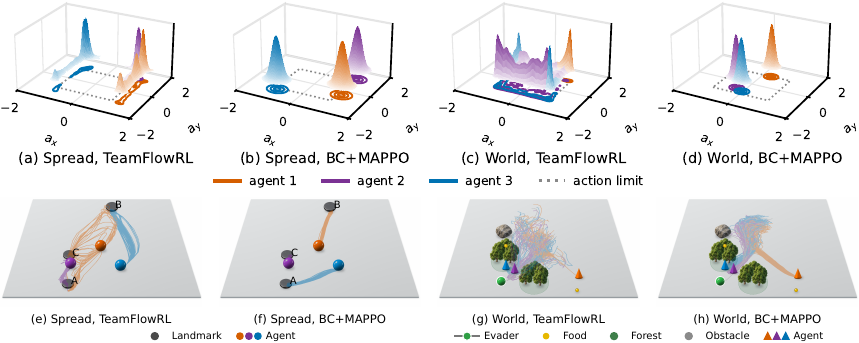}
\caption{Action distributions and MPE trajectories of TeamFlowRL and BC+MAPPO. (a)--(d) Action distributions of the three agents at the first step of one Spread and one World scene, with colors identifying agents. (e)--(h) Paths of 64 MPE rollouts per method from one initial scene, with the same agent colors. Additional diagnostic details are provided in Appendix~\ref{app:behavior-details}.}\label{fig:stochastic-trajectories}\label{fig:action-distribution}
\end{figure*}
\begin{figure*}[!t]\centering
\includegraphics[width=\textwidth]{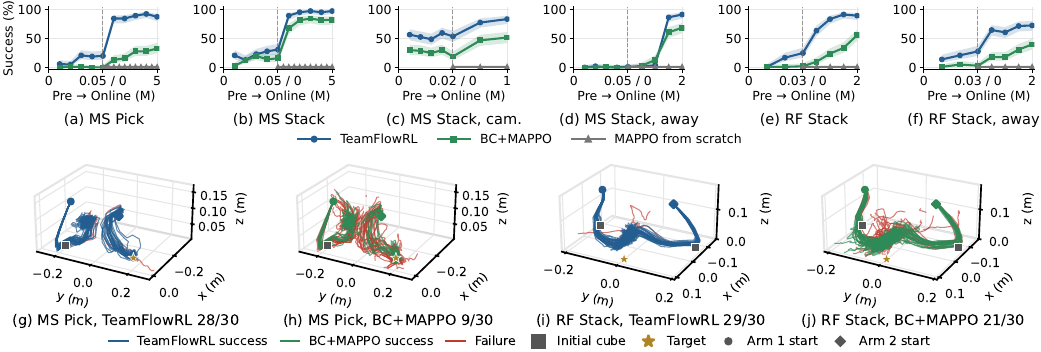}
\caption{Training in multi-arm manipulation. (a)--(f) Success with exploration noise from pretraining through online fine-tuning. The vertical line marks the start of online training, and bands are 95\% intervals over the evaluation episodes. (g)--(j) Tool center point paths of both arms from one initial scene with exploration noise, where red lines are failures.}\label{fig:efficiency}
\end{figure*}
\subsection{Policy Expressiveness}\label{sec:cooperative-behavior}
\textbf{Action distributions.} Figure~\ref{fig:action-distribution}(a)--(d) compares actions generated from the same local input. In the displayed scenes, the fine-tuned flow policy produces separated action modes, whereas BC+MAPPO represents each agent's action with a single Gaussian. The fine-tuned flow policy therefore represents action distributions that the Gaussian policy cannot represent.

\textbf{Cooperative behaviors.} In Spread, three agents must cover three landmarks. We label the landmarks A, B, and C, and write an assignment as the landmarks that agents 1, 2, and 3 end nearest to, so CAB means that agent 1 goes to C, agent 2 to A, and agent 3 to B. A rollout covers all landmarks when each landmark has an agent within distance 0.1. Over 64 stochastic rollouts from one fixed scene, TeamFlowRL covers all landmarks with two different assignments. It uses CAB in 35 rollouts and ACB in 28, as shown in Figure~\ref{fig:stochastic-trajectories}(e)--(f). The remaining rollout leaves one landmark uncovered. We compare with BC+MAPPO, which pretrains a Gaussian policy on the same data and fine-tunes it with MAPPO. BC+MAPPO covers the landmarks in 63 rollouts but always uses the single assignment BCA. In World, Figure~\ref{fig:stochastic-trajectories}(g)--(h) shows that TeamFlowRL also follows more varied pursuit paths than BC+MAPPO from the fixed initial scene. Thus, the fine-tuned flow policy provides both comparable task coverage and
greater behavioral diversity than the unimodal Gaussian baseline. Further behavioral and robot trajectory diagnostics are provided in Appendices~\ref{app:part-experiments} and~\ref{app:part-robot}.

\begin{figure*}[!t]\centering
\includegraphics[width=0.92\textwidth]{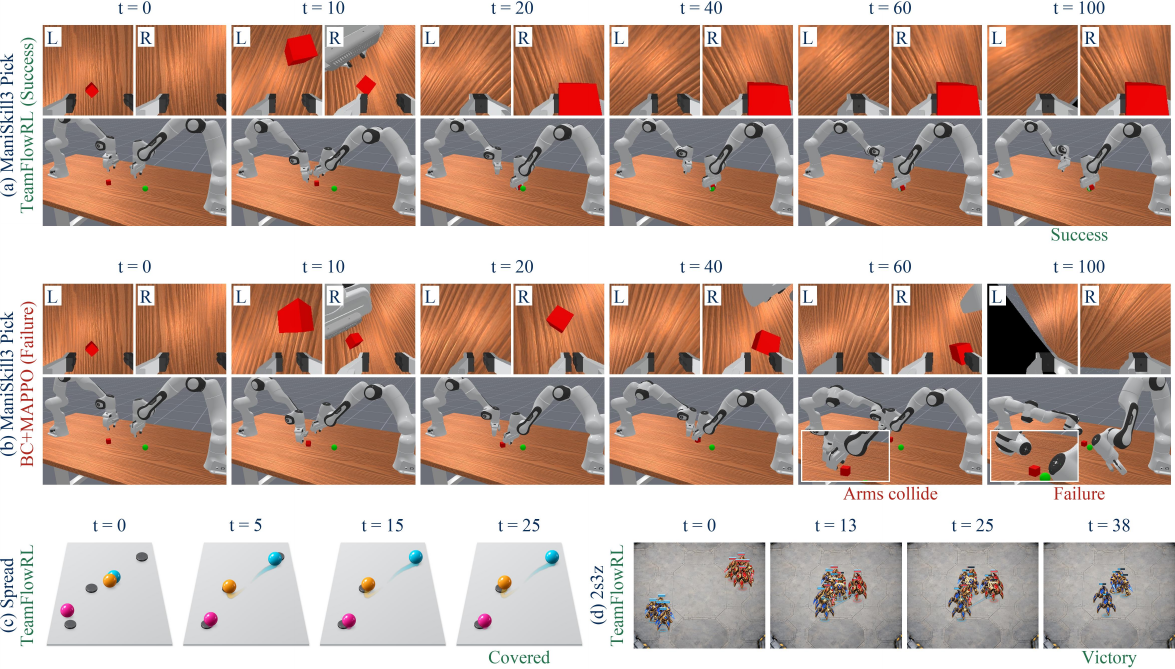}
\caption{Recorded episodes. (a)--(b) TeamFlowRL and BC+MAPPO from the same ManiSkill3 Pick scene under exploration noise, with the left (L) and right (R) wrist views above each frame; insets in (b) enlarge the arms colliding and the cube left on the table. (c)--(d) TeamFlowRL in Spread and in SMAC 2s3z.}\label{fig:rollouts}
\end{figure*}
\subsection{Comparison with Gaussian Policies}\label{sec:bimanual}
We compare TeamFlowRL with BC+MAPPO, which pretrains a one-step Gaussian policy on the same data and fine-tunes it with MAPPO, and with MAPPO trained from scratch.

\textbf{Online adaptation.} Figure~\ref{fig:efficiency}(a)--(f) shows the
training curves of the six robot tasks. After 1 million online steps, the
success of TeamFlowRL with exploration noise averages 68.0\% over the six
tasks, against 31.3\% for BC+MAPPO. Table~\ref{tab:robot-results} shows that
MAPPO trained from scratch succeeds in none of the six tasks at the end of
training. This comparison shows that offline flow pretraining substantially improves
online adaptation over both Gaussian fine-tuning and training MAPPO from
scratch.

\textbf{Performance under exploration noise.} Online training executes each
policy with exploration noise, so we evaluate the success of these executions.
At the end of training, TeamFlowRL has higher success than BC+MAPPO in all six
tasks, by 33.1\% on average, as reported in Table~\ref{tab:robot-results}.
From one fixed initial state per task, 95\% of the TeamFlowRL executions
succeed, against 50\% for BC+MAPPO, as shown in Figure~\ref{fig:efficiency}(g)--(j).
Figure~\ref{fig:rollouts}(a)--(b) shows one such episode: TeamFlowRL lifts the
cube to the goal, whereas BC+MAPPO leaves it on the table. Overall, TeamFlowRL retains its advantage when the policies are evaluated
under the same stochastic execution rule used during online training. These main results evaluate the stochastic training policy. Additional results under each method's deployment rule are provided in Appendix~\ref{app:part-robot}.

\input{tables/robot_results}

\subsection{Ablation Study}
We compare the complete denoising trajectory ratio with per-step clipping and
ratio averaging with clamping on three representative configurations. The
trajectory ratio achieves the best final performance in two of the three
configurations and is never substantially worse than the alternatives. This
result supports using the complete trajectory ratio as the default update rule,
while detailed curves and configuration results are provided in
Appendix~\ref{app:ratio-ablation}.

%% file: tables/robot_results.tex
\begin{table}[!t]
\centering
\caption{Robot success under stochastic execution; bold marks the best final result.}
\label{tab:robot-results}
\begingroup\footnotesize
\setlength{\tabcolsep}{0pt}\renewcommand{\arraystretch}{0.85}
\begin{tabular*}{\linewidth}{@{\extracolsep{\fill}}l@{\hspace{4pt}}ccccc@{}}
\toprule
& MAPPO & \multicolumn{2}{c}{BC+MAPPO} & \multicolumn{2}{c}{TeamFlowRL}\\
\cmidrule(lr){3-4}\cmidrule(lr){5-6}
Task & Trained & Pretrained & Fine-tuned & Pretrained & Fine-tuned\\
\midrule
MS Pick & 0.0\% & 0\% & 34.4\% & 20\% & \textbf{91.6\%}\\
MS Stack & 0.0\% & 16\% & 82.6\% & 31\% & \textbf{95.8\%}\\
MS Stack, cam. & 0.0\% & 19\% & 52.0\% & 54\% & \textbf{84.0\%}\\
MS Stack, away & 0.0\% & 0\% & 69.0\% & 1\% & \textbf{91.6\%}\\
RF Stack & 0.0\% & 2\% & 49.8\% & 25\% & \textbf{90.4\%}\\
RF Stack, away & 0.0\% & 3\% & 40.0\% & 28\% & \textbf{73.0\%}\\
\cmidrule(lr){1-6}
Average & 0.0\% & 6.7\% & 54.6\% & 26.5\% & \textbf{87.7\%}\\
\bottomrule
\end{tabular*}
\par\vspace{1pt}\raggedright\scriptsize
MS/RF: ManiSkill3/RoboFactory; cam.: camera; away: hands-away.
\par\endgroup
\end{table}

%% file: sections/conclusion.tex
\section{Conclusion}
We introduced TeamFlowRL, a two-stage framework that pretrains multi-agent flow policies offline and fine-tunes them through online interaction. TeamFlowRL formulates team policy optimization over decentralized denoising trajectories, using a shared team advantage and separate clipping of each agent's complete trajectory ratio to align local updates with the common team return. Across 51 configurations in seven simulation groups, online fine-tuning improves every offline initialization, with a 47.3\% average gain in normalized score. The fine-tuned flow policies also represent multimodal action distributions and diverse cooperative behaviors, and achieve 33.1\% higher average success under exploration noise than Gaussian-policy fine-tuning across six bimanual tasks.

\paragraph{Limitations.} Our manipulation evaluation is limited to simulated
two-arm cube manipulation. Camera observations are used in only two tasks, and
we do not evaluate physical robots. We use compact, task-specific policy
networks and have not evaluated TeamFlowRL with large VLA models. Finally,
offline data quality affects final online performance differently across tasks.
We have not identified why lower-quality initializations catch up with
higher-quality ones in some tasks but not others, as shown in
Figure~\ref{fig:learning}.

%% file: sections/appendix.tex
\section{Method Details and Theoretical Analysis}\label{app:formulas}\label{app:method-details}
\begingroup
\subsection{Notation and Index Conventions}\label{app:notation}
Environment time $t$, flow time $\tau$, denoising step $k$, and the agent index $i$ describe different
operations. Agent indices are superscripts and environment time is a subscript, as in $h_t^i$. Within one
denoising trajectory the time index is omitted, so $x^{i,k}$ is the variable of agent $i$ after $k$ of
the $K=10$ denoising steps. Denoising step $k$ corresponds to flow time $\tau=k/K$, at which the velocity network is evaluated in Equation~\ref{eq:transition}. A subscript $\tau$, as in $x_\tau^i$, always denotes
flow time and never environment time. Bold symbols collect all $N$ agents.
Table~\ref{tab:notation} lists the symbols.
\input{tables/notation}

\subsection{Offline Flow Objective and Networks}\label{app:offline}
For a dataset pair $(h^i,a^i)$, a latent $z^i\sim\N(0,I)$, and $\tau\sim\mathcal U[0,1]$, the straight
interpolation $x_\tau^i=(1-\tau)z^i+\tau a^i$ has the conditional velocity
\begin{equation}
u(x_\tau^i\mid z^i,a^i)=\frac{d x_\tau^i}{d\tau}=a^i-z^i,
\end{equation}
and conditional flow matching regresses the network onto it, which gives Equation~\ref{eq:fm}. For the
true marginal velocity $u^\star(x,\tau\mid h)=\E[a-z\mid x_\tau=x,h]$, the regression target satisfies
\begin{equation}
\begin{aligned}
&\E\|v_\theta(h,x_\tau,\tau)-(a-z)\|^2\\
&\quad=\E\|v_\theta(h,x_\tau,\tau)-u^\star(x_\tau,\tau\mid h)\|^2+C,
\end{aligned}
\end{equation}
where $C$ does not depend on $\theta$. This is the standard argument that the conditional loss has the
same minimizer as regression onto $u^\star$. The offline teacher generates an action with $K$
deterministic Euler steps,
\begin{equation}
x^{i,k+1}=x^{i,k}+K^{-1}v_\theta(h^i,x^{i,k},k/K),
\end{equation}
for $k=0,\ldots,K-1$, which is the transition mean of Equation~\ref{eq:transition}. The velocity network
is a multilayer perceptron with GELU activations on the concatenation of $h^i$, $x^{i,k}$, and either
$k/K$ or sinusoidal features of $k/K$. The widths and time features are imported with the offline
checkpoint. The noise network $f_\phi$ is a separate perceptron with one hidden layer of width 64 and a
tanh activation on $(h^i,x^{i,k},k/K)$. Its output layer starts with zero weights and a constant bias, so
every coordinate starts at the configured standard deviation:
\begin{equation}
\begin{aligned}\sigma_\phi(h^i,x^{i,k},k)={}&\sigma_{\min}+(\sigma_{\max}-\sigma_{\min})\\&\cdot\operatorname{sigmoid}\bigl(f_\phi(h^i,x^{i,k},k/K)\bigr),\end{aligned}
\end{equation}
with $\sigma_{\min}=0.01$, $\sigma_{\max}=0.5$, and initial value $0.1$.

\subsection{Denoising Trajectory Likelihood}\label{app:path-likelihood}
Given $x^{i,k}$, the mean $\mu_\theta^{i,k}$ and the standard deviation $\sigma_\phi^{i,k}$ in
Equation~\ref{eq:transition} are fixed, and
$x^{i,k+1}=\mu_\theta^{i,k}+\sigma_\phi^{i,k}\odot\epsilon^{i,k}$ is an affine function of
$\epsilon^{i,k}\sim\N(0,I)$. An affine function of a standard Gaussian vector is Gaussian, so
$x^{i,k+1}\mid x^{i,k}\sim\N\bigl(\mu_\theta^{i,k},\operatorname{diag}((\sigma_\phi^{i,k})^2)\bigr)$.
Because each step depends on earlier steps only through $x^{i,k}$, the trajectory probability in
Equation~\ref{eq:path} is the product of these transitions. Each transition is a diagonal Gaussian over
the $d$ action coordinates, so
\begin{equation}
\begin{aligned}
&\log\N\bigl(x^{i,k+1}\mid\mu_\theta^{i,k},\operatorname{diag}((\sigma_\phi^{i,k})^2)\bigr)\\
&\quad=-\sum_{c=1}^{d}\Bigl[\frac{(x^{i,k+1}_c-\mu^{i,k}_{\theta,c})^2}{2(\sigma^{i,k}_{\phi,c})^2}+\log\sigma^{i,k}_{\phi,c}+\tfrac12\log2\pi\Bigr],
\end{aligned}
\end{equation}
and the stored log likelihood is the sum over all steps and coordinates,
\begin{equation}
\ell_{\theta,\phi}(\xi^i\mid h^i)=\sum_{k=0}^{K-1}\log\N\bigl(x^{i,k+1}\mid\mu_\theta^{i,k},\operatorname{diag}((\sigma_\phi^{i,k})^2)\bigr).
\end{equation}
The ratio in Equation~\ref{eq:ratio} is
\begin{equation*}
\rho_t^i=\exp\bigl(\ell_{\theta,\phi}(\xi_t^i\mid h_t^i)-\ell_{\mathrm{old}}(\xi_t^i\mid h_t^i)\bigr).
\end{equation*}
The stored denoising trajectory is held fixed while
$\mu$ and $\sigma$ are recomputed at its prefixes. The marginal probability of the final variable is
represented by
\begin{equation}
q^K_{\theta,\phi}(x^{i,K}\mid h^i)=\int q_{\theta,\phi}(\xi^i\mid h^i)\,dx^{i,0}\cdots dx^{i,K-1},
\end{equation}
which cannot be evaluated directly. The executed action distribution is its pushforward under $g$, with
point masses at the clipping bounds for continuous control.

\begin{proposition}[Trajectory ratios and marginal ratios]\label{prop:path-marginal}
Fix $h^i$ and let $q=q_{\theta,\phi}(\cdot\mid h^i)$ and $q_{\mathrm{old}}$ be the joint probabilities of
two denoising processes of the form of Equation~\ref{eq:path} with strictly positive standard deviations.
Then, for almost every $x^{i,K}$,
\begin{equation}
\E_{\xi^i\sim q_{\mathrm{old}}}\!\left[\frac{q(\xi^i)}{q_{\mathrm{old}}(\xi^i)}\,\middle|\,x^{i,K}\right]=\frac{q^K(x^{i,K})}{q^K_{\mathrm{old}}(x^{i,K})}.
\label{eq:path-marginal}
\end{equation}
If, in addition, differentiation can pass under the integral, then
\begin{equation}
\nabla\log q^K(x^{i,K})=\E_{\xi^i\sim q}\bigl[\nabla\log q(\xi^i)\,\big|\,x^{i,K}\bigr].
\label{eq:fisher}
\end{equation}
\end{proposition}
\begin{proof}
Gaussian transitions with positive variance make $q$ and $q_{\mathrm{old}}$ positive everywhere. Write
$y=(x^{i,0},\ldots,x^{i,K-1})$. The conditional probability of $y$ given $x^{i,K}$ under
$q_{\mathrm{old}}$ is $q_{\mathrm{old}}(\xi^i)/\allowbreak q^K_{\mathrm{old}}(x^{i,K})$, so the left side of
Equation~\ref{eq:path-marginal} equals
\[
\begin{aligned}\int\frac{q_{\mathrm{old}}(\xi^i)}{q^K_{\mathrm{old}}(x^{i,K})}\frac{q(\xi^i)}{q_{\mathrm{old}}(\xi^i)}\,dy&=\frac{1}{q^K_{\mathrm{old}}(x^{i,K})}\int q(\xi^i)\,dy\\&=\frac{q^K(x^{i,K})}{q^K_{\mathrm{old}}(x^{i,K})}.\end{aligned}
\]
For Equation~\ref{eq:fisher},
\begin{equation*}
\nabla q^K(x^{i,K})=\int\nabla q(\xi^i)\,dy=\int q(\xi^i)\,\nabla\log q(\xi^i)\,dy.
\end{equation*}
Dividing by $q^K(x^{i,K})$ and using the conditional probability of $y$ under $q$ gives
the claim.
\end{proof}
The proposition explains why the trajectory ratio is a usable substitute for the intractable ratio of
action marginals: conditioned on the final variable, it is unbiased for the marginal ratio, and its
expected score equals the marginal score. By the law of total variance, the unconditional variance of the trajectory ratio is at least that of the marginal ratio, because the intermediate variables add conditional randomness.

\subsection{Theoretical Analysis}\label{app:proofs}
The analysis concerns the joint \emph{sampling} policy with transition noise,
\begin{equation*}
M_{\theta,\phi}(\boldsymbol\xi\mid\chi)=\prod_iq_{\theta,\phi}(\xi^i\mid h^i),
\end{equation*}
where $\chi$ is a Markov
information state containing the joint histories and each $h^i$ is a fixed projection of $\chi$. The
environment sees only $\mathbf a=(g(x^{1,K}),\ldots,g(x^{N,K}))$. Assume $0<\gamma<1$, bounded rewards,
initial and transition distributions independent of the parameters, fixed legal masks at each history,
and the regularity needed to interchange differentiation and expectation. Let $d_M$ be the normalized
discounted occupancy of $\chi$ and $A^M(\chi,\boldsymbol\xi)$ the exact team advantage, which depends on
$\boldsymbol\xi$ only through $\mathbf a$. The statements below concern the training policy with transition noise. 

\begin{proposition}[Raw denoising trajectories under a fixed action map]\label{prop:clip}
The gradient of the expected return of the sampling policy can be written with the trajectory score
$\nabla\log q_{\theta,\phi}(\xi^i\mid h^i)$. For two policies over denoising trajectories $q,p$ at the same history,
\begin{equation*}
\KL(g_\#q^K\|g_\#p^K)\le\KL(q^K\|p^K)\le\KL(q\|p)
\end{equation*}
whenever the right side is finite.
\end{proposition}
\begin{proof}
For a single decision with return functional $\mathcal R(g(x^{i,K}))$ independent of the parameters,
\begin{equation*}
\nabla\E_{\xi\sim q}\bigl[\mathcal R(g(x^{i,K}))\bigr]=\int\mathcal R(g(x^{i,K}))\,\nabla q(\xi)\,d\xi=\E_{q}\bigl[\mathcal R\,\nabla\log q(\xi)\bigr].
\end{equation*}
Applying this to finite trajectories and taking the
discounted limit gives the score form. Both inequalities follow from the chain rule for relative entropy.
The KL of a joint distribution equals the KL of a marginal plus the expected KL of the conditionals,
which is nonnegative. The first applies it to $(x^{i,K},g(x^{i,K}))$, and the second to
$(\xi^i,x^{i,K})$.
\end{proof}

\Needspace{8\baselineskip}
\begin{proposition}[Factorized joint change]\label{prop:joint}
At a fixed information state, for product policies over denoising trajectories $M=\prod_iq_i$ and
$M'=\prod_ip_i$ with finite individual KL divergences,
\begin{equation*}
\KL(M\|M')=\sum_i\KL(q_i\|p_i),\qquad\prod_i\rho^i=\frac{M'}{M}.
\end{equation*}
\end{proposition}
\begin{proof}
\begin{equation*}
\KL(M\|M')=\E_{\boldsymbol\xi\sim M}\Bigl[\sum_i\log\frac{q_i(\xi^i)}{p_i(\xi^i)}\Bigr]=\sum_i\E_{\xi^i\sim q_i}\Bigl[\log\frac{q_i}{p_i}\Bigr]
\end{equation*}
because each term depends only on $\xi^i$, whose marginal under $M$ is $q_i$. The
ratio identity is the factorization itself.
\end{proof}
The factorization uses independent latents and transition noise across agents. Shared parameters do not
affect it. A bound $\delta$ on the KL averaged over active agents gives at most $N\delta$ for the joint
KL at that history. 

\begin{proposition}[Team policy gradient with shared parameters]\label{prop:team-gradient}
Under the stated assumptions,
\begin{equation}
\begin{aligned}\nabla J={}&\frac{1}{1-\gamma}\E_{\chi\sim d_M,\boldsymbol\xi\sim M}\Bigl[A^M(\chi,\boldsymbol\xi)\\&\cdot\sum_{i=1}^N\nabla\log q_{\theta,\phi}(\xi^i\mid h^i)\Bigr],\end{aligned}
\label{eq:team-gradient}
\end{equation}
where the gradient is taken with respect to $(\theta,\phi)$.
\end{proposition}
\begin{proof}
The log likelihood of a finite trajectory contains $\sum_t\sum_i\log q_{\theta,\phi}(\xi_t^i\mid h_t^i)$.
The environment terms do not depend on the parameters. The score identity, removal of rewards that
precede each score by conditional expectation, replacement of the remaining return by the action value,
and subtraction of the state value, which does not depend on the action, give
Equation~\ref{eq:team-gradient} after the discounted limit. With shared parameters, the product rule sums
the scores of all agents.
\end{proof}
A common team advantage is therefore compatible with decentralized sampling. The implementation replaces
$A^M$ with normalized GAE from a learned critic and averages uniformly over rollout samples.

\begin{proposition}[First-order consistency of agent ratios]\label{prop:first-order}
Fix the collection parameters, write $A^{\mathrm{old}}$ for the corresponding advantage, and define, with
expectations under the collection occupancy and policy,
\begin{equation}
\begin{aligned}
S_{\mathrm{agent}}&=\E\Bigl[A^{\mathrm{old}}\sum_i(\rho^i-1)\Bigr],\\
S_{\mathrm{joint}}&=\E\Bigl[A^{\mathrm{old}}\Bigl(\prod_i\rho^i-1\Bigr)\Bigr].
\end{aligned}
\end{equation}
At the collection parameters both vanish, their gradients are equal to $(1-\gamma)\nabla J$, and
replacing each $\rho^iA^{\mathrm{old}}$ by its clipped counterpart leaves the gradient unchanged.
\end{proposition}
\begin{proof}
Every ratio equals one at the collection parameters, so there
\begin{equation*}
\nabla\prod_i\rho^i=\sum_i\nabla\rho^i=\sum_i\nabla\log q_{\theta,\phi}(\xi^i\mid h^i).
\end{equation*}
Substituting into the definitions and using Proposition~\ref{prop:team-gradient} gives the gradient
identity. The clipping interval contains one in its interior, so the clipped and unclipped terms agree in
a neighborhood and have the same derivative.
\end{proof}
The agreement is local: after the first update of an iteration, sums and products of ratios differ.
 By
Proposition~\ref{prop:path-marginal}, the same local agreement holds between the trajectory ratio and the
ratio of action marginals.

\subsection{Comparison with the Gaussian Baseline}\label{app:compare-gaussian}
The baseline BC+MAPPO fine-tunes a one-step Gaussian policy with MAPPO. Table~\ref{tab:compare-gaussian}
lists where the two optimization problems differ. In the analysis above, the statements that hold for any
product of agent policies are stated for policies over denoising trajectories.
Proposition~\ref{prop:path-marginal} and the trajectory form of Proposition~\ref{prop:clip} have no
counterpart for a one-step Gaussian, whose action likelihood can be evaluated directly. The one-step
baseline is deployed with its latent fixed at zero, whereas TeamFlowRL samples a fresh latent at every
step during both training and deployment.

\begin{table*}[!hbp]\centering\small
\caption{TeamFlowRL and the Gaussian baseline BC+MAPPO. Values for BC+MAPPO follow its configuration in the robot tasks, listed in Table~\ref{tab:robot-configs}.}\label{tab:compare-gaussian}
\begin{tabularx}{\textwidth}{@{}lXX@{}}
\toprule
 & TeamFlowRL & BC+MAPPO\\\midrule
Initial policy & Ten-step flow teacher $v_\theta$ from flow matching & One-step student network, latent fixed at zero\\
Sampling distribution & Markov chain of $K=10$ Gaussian transitions with learnable noise $\sigma_\phi(h^i,x^{i,k},k)\in[0.01,0.5]$ & Gaussian around the student output with a learned standard deviation shared across states and agents\\
Stored sample & Complete denoising trajectory $\xi_t^i=(x_t^{i,0},\ldots,x_t^{i,K})$ & Raw action before clipping\\
Likelihood & Sum of $K$ transition log probabilities (Equation~\ref{eq:path}). The action marginal cannot be evaluated directly & Gaussian likelihood of the raw action\\
Ratio & One trajectory ratio per agent (Equation~\ref{eq:ratio}) & One action ratio per agent\\
Clipping & Per-agent ratio, $\varepsilon=0.2$ by default, computed in log space & Per-agent ratio, $\varepsilon=0.05$ in the robot tasks\\
Regularization & None in reported runs ($\beta=0$, no entropy term) & Reference KL coefficient 0.01 and entropy coefficient 0.001 in the robot tasks\\
Update-size control & Conditional trajectory KL summed over $K$ transitions, threshold 0.02, optional backtracking & Analytic Gaussian KL of the action, threshold 0.02\\
Deployment & Random initial latent, deterministic Euler steps, so the action stays stochastic & Mean action with the latent fixed at zero\\
\bottomrule
\end{tabularx}
\end{table*}
\endgroup

\section{Implementation Details}\label{app:part-other}
\subsection{Inputs, Action Map, and Update Control}\label{app:impl}
\paragraph{Inputs and initialization.} The actor conditions only on its local input. Continuous control
uses local observation features with the identity encoding supplied by the environment adapter. For
StarCraft, the actor conditions on local recurrent features from a pretrained encoder that remains frozen
during online optimization. The online actor imports the teacher velocity network from the offline
checkpoint. 
All reported configurations use $K=10$ denoising steps and $\beta=0$ in Equation~\ref{eq:losses}.
\paragraph{Action map.} The actor stores the sampled denoising trajectory before any action
transformation. The executed action is $a^i=\clip(x^{i,K},-1,1)$ for continuous control and the largest
legal coordinate of $x^{i,K}$ for StarCraft.
\paragraph{Valid trajectory ratio.} Three implementation requirements make the ratio in
Equation~\ref{eq:ratio} valid. First, each denoising trajectory is sampled from the noisy transition
chain of Equation~\ref{eq:transition} itself, and its log likelihood is stored at collection. The update
recomputes the likelihood of the same stored trajectory, so $\rho_t^i=1$ at the collection parameters,
which we check at every update. Second, $\log\rho_t^i$ is the exact sum of the transition log
probabilities over all $K$ steps and action coordinates, without averaging or clamping. Third, each agent
has its own ratio, clipped against the shared team advantage. The
stored trajectory is detached, so gradients enter the velocity and noise networks only through the
conditional means and standard deviations at the stored prefixes.
\paragraph{Update size and execution.} Gradient clipping and a conditional trajectory KL check limit the
update size: once the trajectory KL exceeds its threshold, the remaining actor updates of the iteration
are skipped, and some tasks additionally use KL backtracking, described in Appendix~\ref{app:online-details}. Bounded
transition means define a separate repair variant, described in Appendix~\ref{app:variants}. Training samples with transition noise, and deployment follows the mean update.

\subsection{Online Fine-Tuning Settings}\label{app:online-details}\label{app:variants}
With collection critic parameters $\psi_{\mathrm{old}}$, the implementation computes GAE backward over
each rollout of 128 steps,
\begin{equation}
\begin{aligned}
\delta_t&=r_t+\gamma(1-u_t)V_{\psi_{\mathrm{old}}}(s^{+}_{t+1})-V_{\psi_{\mathrm{old}}}(s_t),\\
\widehat A_t&=\delta_t+\gamma\lambda(1-d_t)\widehat A_{t+1},\qquad \widehat R_t=\widehat A_t+V_{\psi_{\mathrm{old}}}(s_t),
\end{aligned}
\label{eq:gae}
\end{equation}
where $s^{+}_{t+1}$ is the successor state before any reset, $\gamma=0.99$, and $\lambda=0.95$. A time
limit has $u_t=0$ and $d_t=1$, so it bootstraps from the final state but cuts the trace. The value
targets are formed before normalization. The actor advantage is normalized once per rollout, with the
population standard deviation, and copied to every agent:
\begin{equation}
\widetilde A_t=\frac{\widehat A_t-\overline A}{\max\bigl(\operatorname{std}(\widehat A),10^{-8}\bigr)},\qquad \widetilde A_t^i=\widetilde A_t\ \text{for all } i.
\label{eq:normadv}
\end{equation}
The actor and critic minimize
\begin{equation}
\begin{aligned}
\mathcal L_{\mathrm{actor}}&=-L_{\mathrm{clip}},\\
\mathcal L_V&=\tfrac12\,\E_t\bigl[(V_\psi(s_t)-\widehat R_t)^2\bigr],
\end{aligned}
\label{eq:losses}
\end{equation}
where the return target $\widehat R_t$ is held fixed.
The implementation evaluates Equation~\ref{eq:ppo} in log space and clips only the side selected by the
advantage sign,
\begin{equation}
\begin{aligned}
&\min\bigl(\rho\widetilde A,\clip(\rho,1-\varepsilon,1+\varepsilon)\widetilde A\bigr)\\
&\quad=\begin{cases}
e^{\min(\log\rho,\,\log(1+\varepsilon))}\,\widetilde A, & \widetilde A\ge0,\\
e^{\max(\log\rho,\,\log(1-\varepsilon))}\,\widetilde A, & \widetilde A<0,
\end{cases}
\end{aligned}
\end{equation}
which is an identity. For $\widetilde A\ge0$, the minimum of $\rho\widetilde A$ and
$\clip(\rho)\widetilde A$ is $\min(\rho,1+\varepsilon)\widetilde A$. For $\widetilde A<0$, it is
$\max(\rho,1-\varepsilon)\widetilde A$. Update size is measured by the conditional trajectory KL from the
collection policy,
\begin{equation}
\begin{aligned}D_t^i=\sum_{k=0}^{K-1}\KL\bigl(&\N(\mu_{\mathrm{old}}^{i,k},\operatorname{diag}((\sigma_{\mathrm{old}}^{i,k})^2))\\&\|\,\N(\mu_\theta^{i,k},\operatorname{diag}((\sigma_\phi^{i,k})^2))\bigr),\end{aligned}
\end{equation}
evaluated at the stored prefixes and averaged with weights $m_t^i$. By the chain rule for relative
entropy, its expectation over prefixes from the collection policy equals the trajectory KL
$\KL(q_{\mathrm{old}}\|q_{\theta,\phi})$. Without backtracking, after each actor minibatch step the
minibatch value is compared with the target 0.02. If it is exceeded, the value over the whole rollout is
computed. If that also exceeds 0.02, the remaining actor updates of the iteration stop while critic
updates continue. The rollout value is also checked at the end of each epoch. The triggering step is
kept. With backtracking enabled, a step is accepted only if the rollout value is at most the target.
Otherwise, parameters and Adam state are restored, and the step is retried with the learning rate halved,
up to the configured number of retries. The first iteration updates only the critic.
Four configurations use task-specific optimization settings: 2-HalfCheetah Good, 5m\_vs\_6m Medium, Tag Random, and Spread Random. They keep the ten-step generator and the complete trajectory ratio, and their settings are provided in the released code.

The centralized critic is a multilayer perceptron with two 256-unit hidden layers and hyperbolic tangent
activations. The conditional noise network has one 64-unit hidden layer with a hyperbolic tangent
activation. The generator architecture is imported from the source checkpoint. The source checkpoint also determines the local recurrent encoder for StarCraft.
The encoder remains frozen so stored local features can be used consistently when replaying likelihoods.
\subsection{Bounded Repair}
\begin{figure*}[!hbp]\centering
\includegraphics[width=\textwidth]{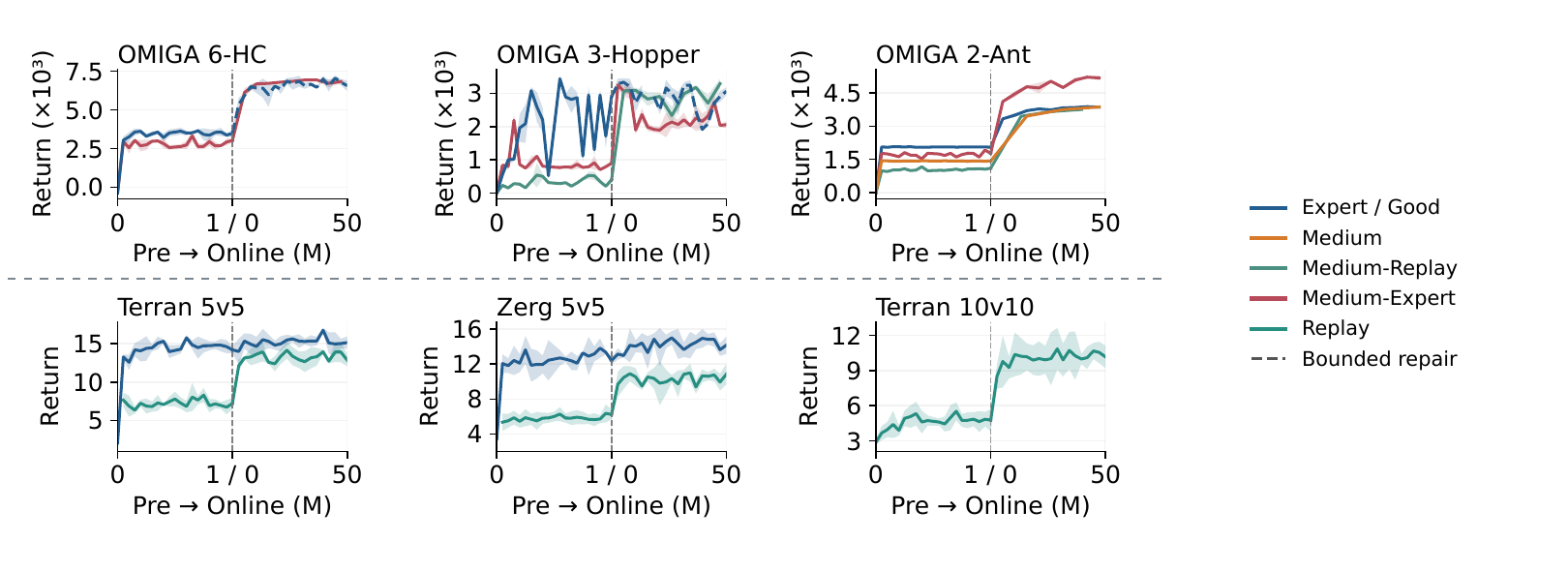}
\caption{Learning curves on OMIGA MuJoCo and SMACv2, drawn as in Figure~\ref{fig:learning}. Dashed curves show runs with bounded repair, and 6-HC denotes 6-HalfCheetah.}\label{fig:learning-appendix}
\end{figure*}

The bounded repair applies $b\tanh(\mu/b)$ to the transition mean $\mu_\theta$ in
Equation~\ref{eq:transition}, with $b=10$. The Gaussian noise is added after this transformation.   Two OMIGA configurations use this
structural change.  The 6-HalfCheetah
Expert and 3-Hopper Expert runs each complete 50 million steps and appear in Table~\ref{tab:omiga-smacv2}
and Figure~\ref{fig:learning-appendix}. Their offline references are the original step zero evaluations
of the same teachers.

\section{Benchmark Experiments}\label{app:part-experiments}
\subsection{Experimental Protocol and Reporting
Conventions}\label{app:results}\label{app:experimental-protocol}\label{app:table-smac}
Table~\ref{tab:hyper} gives the default configuration, and the configuration of every task is provided in the released code. %
Online budgets count joint environment transitions.
\input{tables/online_settings}

Table~\ref{tab:main-results} in the main text reports MPE, OG-MARL, and SMAC, and
Table~\ref{tab:omiga-smacv2} reports OMIGA and SMACv2. Each offline source provides a teacher pretrained
for one million updates. All benchmark entries of TeamFlowRL share an online budget of 50 million
environment steps.  The robot tasks keep their own budgets of 1 to 5 million steps. The OMIGA
Expert entries for 6-HalfCheetah and 3-Hopper use bounded repair at 50 million steps and are marked
$^{\ddagger}$. The learning curves also include the two SMACv2 Good datasets.
\input{tables/omiga_smacv2}

The MPE panel reports normalized scores. For raw return $R$, the score is
\[100\frac{R-R_{\mathrm{random}}}{R_{\mathrm{expert}}-R_{\mathrm{random}}}.\] The remaining
panels report raw team returns. The learning curves and the paired comparisons in the text use raw
returns throughout.

Published baseline values retain their source protocols. SMAC and SMACv2 baselines follow
MAC-Flow~\citep{macflow}, Table 1. MPE baselines follow MAC-Flow, Table 2, except the entries marked
$^{\mathrm m}$ from MADiff~\citep{madiff}, Table 1, and DOM2~\citep{dom2}, Table 1. DOM2 uses its own
collected datasets. Its entries convert the raw returns in DOM2, Table 1, into normalized scores with the
expert and random reference returns given in the DOM2 appendix (p.~22). The DOM2 Spread average is computed by us from its four Spread entries.
OG-MARL baselines follow MADiff, Table 1, OMSD entries follow OMSD, Table 2, and OMIGA baselines follow MAC-Flow, Table 2. 

Entries marked $^{\mathrm r}$ are local reproductions of the offline MAC-Flow student. Their settings are provided in the released code. The three SMACv2 Replay entries marked $^{\mathrm r}$ in Table~\ref{tab:omiga-smacv2} come from the MAC-Flow pretraining run that initializes TeamFlowRL on these tasks. MADiff-D uses decentralized
execution, while MADiff-C uses centralized execution. Published averages retain their source coverage
except for OMIGA, whose baseline averages equally weight the nine reported combinations of task and
dataset. The TeamFlowRL Spread average equally weights four datasets, and its SMACv2 average equally
weights three Replay tasks. 

Light green shading and boldface mark the largest displayed mean. Light gray shading and underlining mark
the second-largest distinct mean. Equal rounded means share ranks. 
M-R and M-E denote Medium-Replay and Medium-Expert. In the main tables, a dagger marks a normalized MPE
score.

\paragraph{Evaluation details.}  
  The offline
flow comparison uses the same evaluation rule as TeamFlowRL.  In MPE, the Medium-Replay and Random Spread initializations also
improve but end below Expert. Table~\ref{tab:continuous-main} reports all MPE combinations on the
normalized score scale used by the published references.
\input{tables/robot_configs}

{\subsection{Trajectory Ratio Ablation}\label{app:ratio-ablation}
We change one choice of the trajectory ratio at a time and keep every other setting of the default
configuration, with a budget of 50 million environment steps. We compare the per-agent trajectory ratio
over the complete denoising trajectory with two alternative constructions: clipping the ratio of each
denoising step separately, as DPPO does for single-agent denoising transitions~\citep{dppo}, and
averaging the trajectory log probability over steps and coordinates before clamping it to $[-5,2]$.
Table~\ref{tab:ratio-ablation} reports the return at 50 million steps and the mean return over the last 5 million steps.

\begin{table}[!hbp]\centering\footnotesize
\caption{Variants of the TeamFlowRL denoising trajectory ratio. Final: return at the end of training. Last 10\%: mean return over the last 10\% of training.}\label{tab:ratio-ablation}
\setlength{\tabcolsep}{4pt}
\begin{tabular}{@{}lcccccc@{}}
\toprule
 & \multicolumn{2}{c}{Spread Medium} & \multicolumn{2}{c}{Spread M-R} & \multicolumn{2}{c}{2-Ant Poor}\\
\cmidrule(lr){2-3}\cmidrule(lr){4-5}\cmidrule(lr){6-7}
Ratio & Final & Last 10\% & Final & Last 10\% & Final & Last 10\%\\\midrule
Complete trajectory per agent (default) & 449.6 & 448.9 & 300.3 & 305.4 & 3195 & 3162\\
Clipped per denoising step & 401.5 & 406.2 & 301.2 & 301.8 & 3195 & 2971\\
Averaged and clamped & 441.2 & 434.7 & 255.5 & 263.2 & 3035 & 3052\\
\bottomrule
\end{tabular}
\end{table}

Clipping each denoising step separately lowers the return in Spread Medium by about 43 and the late mean
in 2-Ant Poor by about 190, and leaves Spread Medium-Replay unchanged. Averaging and clamping the log
probability changes little: although the clamp is active for about 74\% of the samples at the end of
training, the learnable transition noise keeps enough samples inside the clamping range. With a fixed and
smaller transition noise, the averaged log probability starts above the upper bound and most gradients
vanish, so the clamp is harmful only in combination with that noise.}

\subsection{Behavioral Diagnostic Details}\label{app:behavior-details}
The diagnostic uses the Expert checkpoints of TeamFlowRL and BC+MAPPO and the MAPPO checkpoint trained
from scratch. The stochastic Gaussian baselines sample their learned action distributions. 
\begin{figure*}[!hbp]\centering
\includegraphics[width=\textwidth]{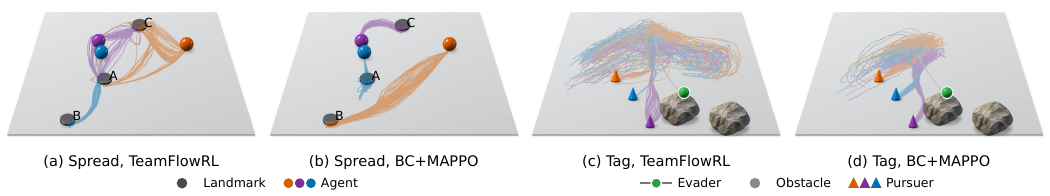}
\caption{MPE rollouts drawn as in Figure~\ref{fig:stochastic-trajectories}, with 64 rollouts per method from one initial scene. (a)--(b) Spread from a second initial scene. (c)--(d) Tag, where colored paths show the three pursuers and gray paths show the evader.}\label{fig:mpe-topview-appendix}
\end{figure*}

The diagnostic uses 64 rollouts from each fixed initial scene. Assignment labels describe the nearest
landmark, while full coverage requires every landmark to be within the stated distance threshold. 

\renewcommand{\topfraction}{0.95}\renewcommand{\bottomfraction}{0.9}\renewcommand{\textfraction}{0.05}\renewcommand{\floatpagefraction}{0.9}\setcounter{topnumber}{4}\setcounter{totalnumber}{6}\renewcommand{\dbltopfraction}{0.95}\renewcommand{\dblfloatpagefraction}{0.9}\setcounter{dbltopnumber}{4}
\section{Robot Manipulation Experiments}\label{app:part-robot}
\subsection{Robot Manipulation Setup}\label{app:robot-setup}
\paragraph{ManiSkill3 tasks.} In TwoRobotPickCube, the cube starts within reach of the left arm only, and
the goal position lies within reach of the right arm only, so the arms must pass the cube between them to
lift it to the goal. An episode succeeds when the cube is within 0.025\,m of the goal and the right arm
is static. In TwoRobotStackCube, each arm can reach only one of two cubes. The right arm places its cube
in a goal region on the midline between the robots, and the left arm stacks the second cube on top. An
episode succeeds when the stacked cubes rest in the goal region and neither arm holds a cube. Each arm
observes its own joint positions and velocities, its own end effector pose, the poses of the cubes, the
goal position, the relative positions between these objects, the remaining time, and an agent indicator,
but not the other arm. Object and goal poses are privileged information, because a physical robot would
have to estimate them from its sensors, so the state results measure cooperative control with known object poses, and Appendix~\ref{app:maniskill-visual} replaces these poses with camera observations. 

\paragraph{Pretraining and online training.} Offline pretraining uses all successful demonstrations
released with ManiSkill3, which number 983 for TwoRobotPickCube and 1007 for TwoRobotStackCube, split
into 90\% for training and 10\% for validation. Both methods start from the same pretraining run of
50,000 updates, in which a frozen recurrent encoder summarizes the observation history of each arm.
TeamFlowRL starts from the ten-step flow policy. BC+MAPPO starts from a one-step policy distilled from
the flow policy, with the latent fixed at zero so that the output is the mean of its Gaussian policy.
Online fine-tuning runs for 5 million environment steps with 64 parallel environments, and each episode
lasts 100 environment steps.    

\paragraph{RoboFactory.} In TwoRobotsStackCube from RoboFactory, each arm receives its own joint and end
effector state, the cube poses, the goal region, and an agent indicator, but not the other arm. Each
action sets joint position changes for one environment step, and each episode lasts 100 steps. Training
uses a staged dense reward that we define for reaching, grasping, lifting, aligning, and placing,
normalized to a maximum of one, and success follows the evaluation function of RoboFactory. The demonstrations come from a PPO expert trained on the dense reward. Pretraining on the scripted motion planning demonstrations provided with RoboFactory gives 0\% closed-loop success. Both methods start from the same pretrained checkpoint. Online fine-tuning uses the
optimization settings of the ManiSkill state experiment for 2 million environment steps. 

\paragraph{Release and withdrawal.} The main table and the RoboFactory learning curve report hands-away success, which additionally requires both arms to release the blocks and withdraw while the stack stands on its own. Appendix~\ref{sec:release_withdraw} defines this variant.

\subsection{Success under the Deployment Rule}\label{app:deploy-success}
\begin{figure}[!htbp]\centering
\includegraphics[width=\textwidth]{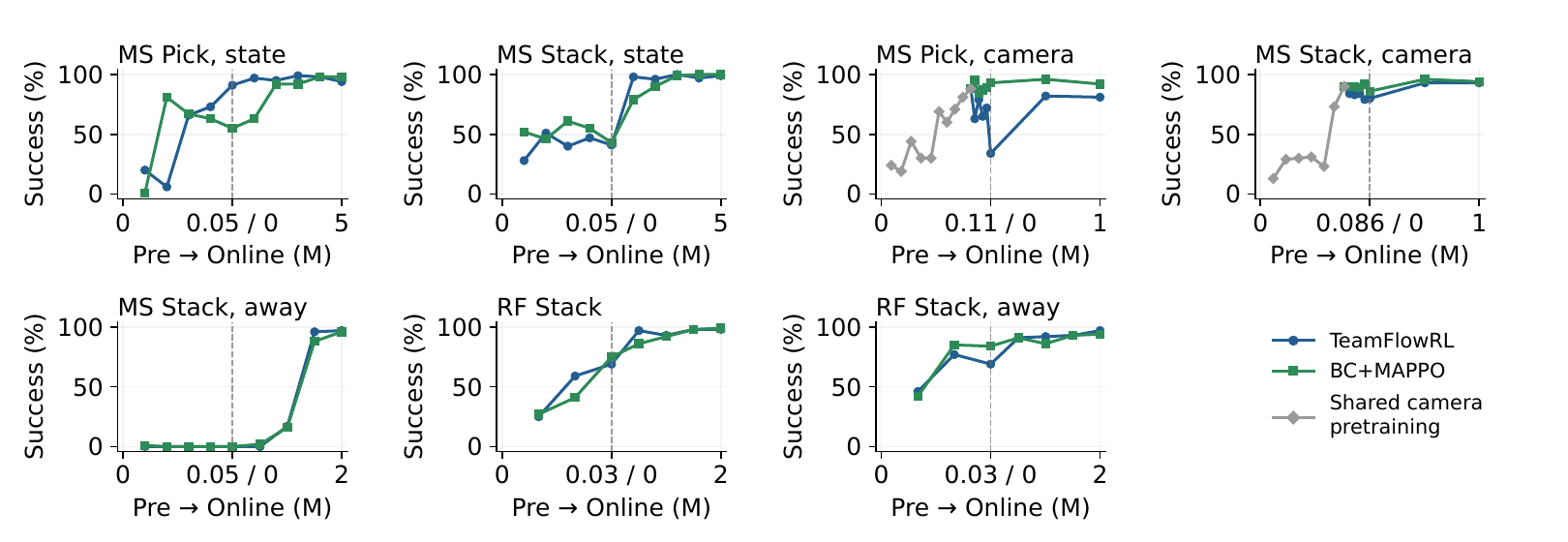}
\caption{Success under each method's deployment rule from pretraining through online fine-tuning, with 100 episodes per point. The horizontal axis is drawn as in Figure~\ref{fig:learning}, with the pretraining length marked at the start of online training. In the camera panels, grey points show the flow pretraining and DAgger shared by both methods, after which each method fits its own action head on the frozen encoder. ``Away'' denotes the hands-away variant.}\label{fig:deploy-curves}
\end{figure}

Figure~\ref{fig:deploy-curves} shows the success of both methods under their deployment rules from
pretraining through online fine-tuning in the seven robot tasks, evaluated on 100 episodes per point.
TeamFlowRL draws new initial latents and adds no transition noise, and BC+MAPPO executes its Gaussian mean.
At the end of online training, the success rates of the two methods differ by at most 5\% in every task. Although the final success is close, TeamFlowRL keeps the multimodal action distribution of the flow policy, because each execution draws new initial latents, whereas BC+MAPPO executes a single mean action. Appendix~\ref{app:deploy-trajectories} shows the resulting variation of the executed paths. Under
the stochastic execution of Table~\ref{tab:robot-results}, TeamFlowRL averages 87.7\% over the six tasks
and BC+MAPPO 54.6\%.

\ifnum\maniskillHasComparison=1
\subsection{Trajectories under the Deployment Rule}\label{app:deploy-trajectories}
Figure~\ref{fig:deploy-trajectories} shows the tool center point paths of both arms when each policy is
run 31 times from one initial scene under its deployment rule. TeamFlowRL draws new initial latents in
every run and adds no transition noise, and BC+MAPPO executes its Gaussian mean. To separate the
variation caused by the simulator, we also run TeamFlowRL with the same latents in all 31 runs. We
measure the spread of the paths as the root mean square distance of the tool center points from their
mean over the runs, over the first 40 control steps and both arms, and average it over four MS Pick
scenes and eight RF Stack scenes. On MS Pick, the spread is 1.6 cm for TeamFlowRL, 0.4 cm with the same latents, and 0.3 cm for BC+MAPPO. On RF Stack, it is 0.9 cm, 0.2 cm, and 0.3 cm. The latent sampling of TeamFlowRL
varies the executed path several times more than the simulator alone, whereas BC+MAPPO follows a single
path. The variation stays at the centimeter level around one route. At least 98\% of the runs succeed
in every group.

\begin{figure}[!htbp]\centering
\includegraphics[width=0.75\textwidth]{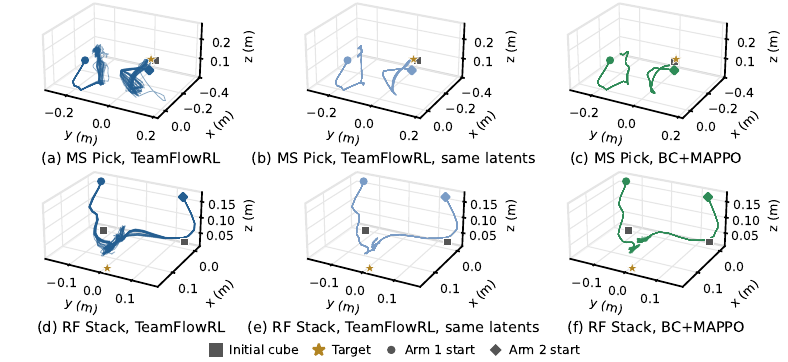}
\caption{Tool center point paths under the deployment rule, with 31 runs from one initial scene per task. (a), (d) TeamFlowRL with new initial latents in every run. (b), (e) TeamFlowRL with the same latents in all runs. (c), (f) BC+MAPPO. Axes are shared within each row.}\label{fig:deploy-trajectories}
\end{figure}

\subsection{Bimanual Stacking under Exploration Noise}\label{app:maniskill-frames}\label{app:rollouts-state}
Figure~\ref{fig:rollouts-compare} shows TwoRobotStackCube in ManiSkill3 in the layout of Figure~\ref{fig:rollouts}.
TeamFlowRL and BC+MAPPO start from the same scene and both execute with exploration noise. In this scene,
TeamFlowRL succeeds in 93\% of 30 runs and BC+MAPPO in 77\%. In the illustrated TeamFlowRL run, the right
arm places the green cube in the goal region and the left arm stacks the blue cube on it. In the illustrated
BC+MAPPO run, the arms bring both cubes into the goal region, but the cubes end side by side and are not
stacked at the final step.

\begin{figure*}[!t]\centering
\includegraphics[width=0.92\textwidth]{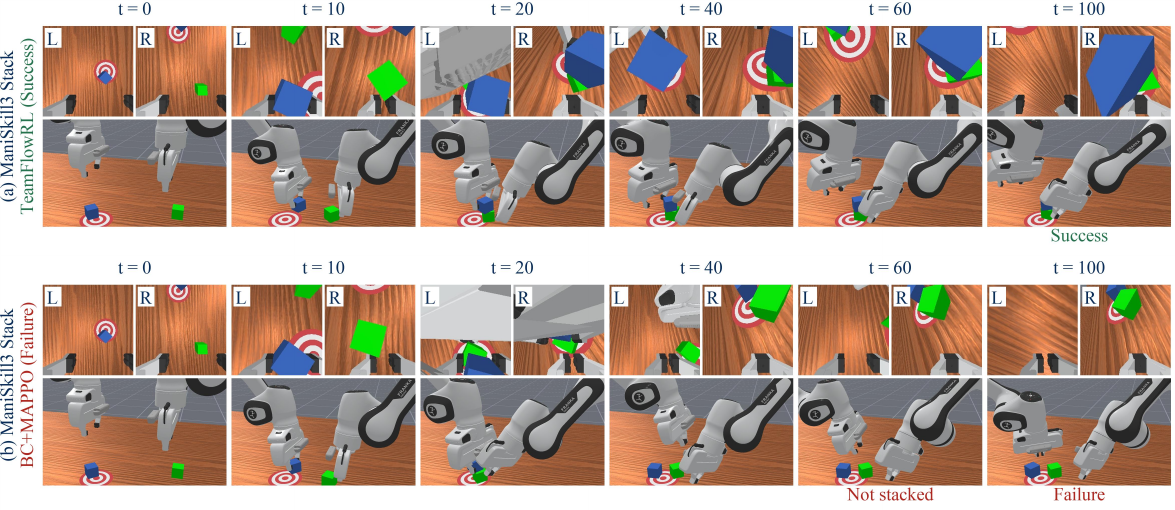}
\caption{Recorded TwoRobotStackCube episodes in ManiSkill3. (a)--(b) TeamFlowRL and BC+MAPPO from the same scene under exploration noise, with the left (L) and right (R) wrist views above each frame.}\label{fig:rollouts-compare}
\end{figure*}
\begin{figure*}[!t]\centering
\includegraphics[width=0.88\textwidth]{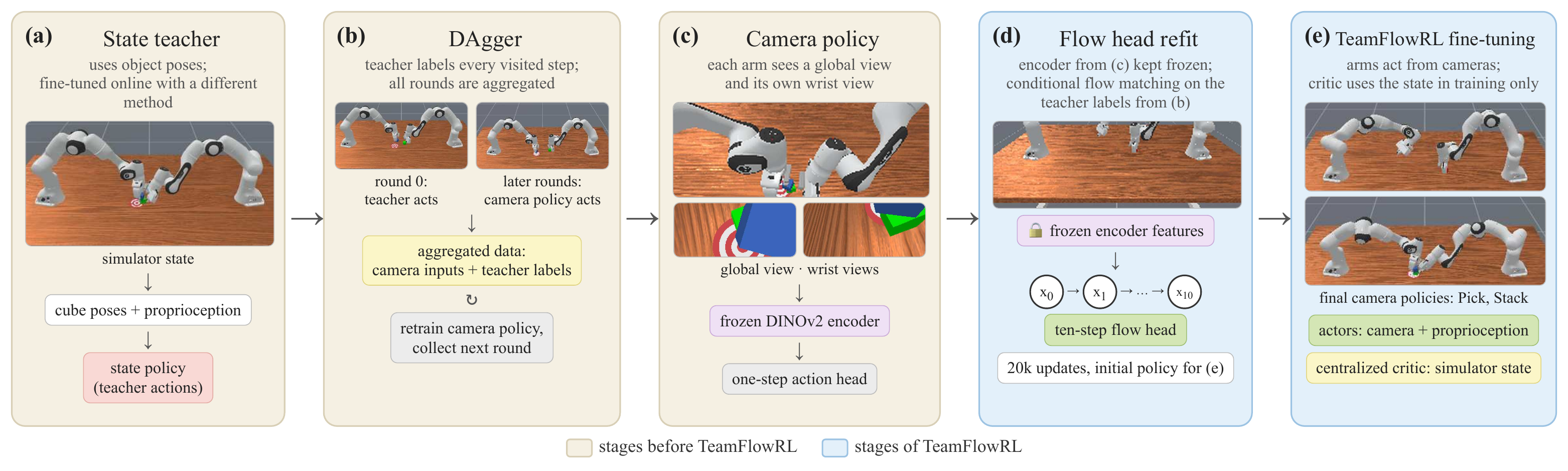}
\caption{Training TeamFlowRL with camera observations through DAgger, flow head fitting on a frozen encoder, and online fine-tuning. Blue panels mark the stages of TeamFlowRL, and beige panels mark the preceding stages.}\label{fig:maniskill-pipeline}
\end{figure*}

\fi

\subsection{Bimanual Manipulation with Camera Observations}\label{app:maniskill-visual}

Figure~\ref{fig:maniskill-pipeline} summarizes how the camera policy for TeamFlowRL is obtained. Each arm receives a global view shared by both arms and its own wrist view, encoded by a frozen DINOv2 model~\citep{dinov2}, together with its proprioceptive state, the goal position, the remaining time, and an agent indicator. The cube poses are available only to the state teacher and to the centralized critic. The state teacher trains a camera policy through DAgger. A camera policy pretrained only on the offline demonstrations reaches 22\% success in TwoRobotPickCube and 23\% in TwoRobotStackCube under the deployment rule, because its own rollouts soon reach states that the demonstrations do not cover. In DAgger, the camera policy controls the arms, the state teacher labels every visited state, and the camera policy is trained on the data of all rounds. The DAgger policy uses a one-step action head, whereas TeamFlowRL optimizes the likelihood of the ten-step denoising trajectory. We therefore freeze the encoder of this policy and fit a ten-step flow head with conditional flow matching on the DAgger data, using the teacher actions as targets. Online fine-tuning keeps the encoders frozen and runs for one million environment steps with the optimization settings of the state experiment. The remaining settings are provided in the released code.

\ifnum\robofactoryFramesReady=1
\fi

\FloatBarrier
\input{release_materials/release_section.tex}

\FloatBarrier

%% file: tables/notation.tex
\begin{table*}[!hbp]
\centering\small\setlength{\tabcolsep}{4pt}
\caption{Symbols used in training, deployment, and analysis.}\label{tab:notation}
\begin{tabular}{@{}p{.24\linewidth}p{.69\linewidth}@{}}
\toprule Symbol & Meaning \\\midrule
\multicolumn{2}{l}{\textbf{Indices, clocks, and local information}}\\
$i,N$ & Agent index and number of controlled agents ($N=|\mathcal I|$).\\
$t,T$ & Environment step and episode horizon.\\
$k,K$ & Denoising step and number of flow steps, with $K=10$.\\
$\tau$ & Continuous interpolation time between latent noise and data.\\
$h_t^i,s_t$ & Local actor input and centralized critic input.\\
$a_t^i,r_t$ & Executed local action and shared team reward.\\
\multicolumn{2}{l}{\textbf{Flow initialization and denoising trajectories}}\\
$z,I$ & Standard Gaussian latent and identity covariance matrix.\\
$x_\tau$ & Interpolated offline flow input.\\
$v_\theta,\theta$ & Conditional velocity field and its shared parameters.\\
$\mathcal L_{\mathrm{FM}}$ & Conditional flow matching objective.\\
$x^{i,k},\xi^i$ & Intermediate denoising variable and complete stored trajectory $(x^{i,0},\ldots,x^{i,K})$.\\
$\mu_\theta$ & Conditional transition mean.\\
$f_\phi,\sigma_\phi,\phi$ & Noise network, conditional transition standard deviation, and its parameters.\\
$\sigma_{\min},\sigma_{\max}$ & Lower and upper bounds on the conditional standard deviation.\\
$\epsilon^{i,k}$ & Independent standard Gaussian transition noise.\\
$p_0,q_{\theta,\phi},M_{\theta,\phi}$ & Initial latent probability, one-agent trajectory probability, and factorized joint trajectory policy.\\
$g,\pi_{\theta,\phi},U$ & Action transformation, induced action distribution, and measurable action set.\\
$b$ & Bound on the transition mean in the repaired variant.\\
\multicolumn{2}{l}{\textbf{Team advantages and online optimization}}\\
$J,\gamma$ & Expected discounted team return and discount factor.\\
$V_\psi,\psi$ & Centralized critic and its parameters.\\
$u_t,d_t$ & True termination indicator and episode boundary indicator.\\
$\delta_t,\lambda,\widehat A_t$ & Temporal-difference residual, GAE trace coefficient, and team advantage estimate.\\
$\ell_{\theta,\phi},\ell_{\mathrm{old}}$ & Sums of conditional transition log probabilities under the current and collection policies.\\
$\rho_t^i,\varepsilon$ & Per-agent likelihood ratio of the denoising trajectory and PPO clipping threshold.\\
$m_t^i,L_{\mathrm{clip}}$ & Actor activity mask and clipped PPO objective.\\
$\E,\KL,\clip$ & Expectation, Kullback--Leibler divergence, and clipping operator.\\
\multicolumn{2}{l}{\textbf{Reported scores}}\\
$R,R_{\mathrm{random}},R_{\mathrm{expert}}$ & Raw episode return and the reference returns for MPE normalization.\\
\bottomrule
\end{tabular}
\end{table*}

%% file: tables/online_settings.tex
\begin{table*}[!hbp]
\centering
\caption{Online fine-tuning settings. Online budgets count joint environment transitions, with one transition per joint environment step.}
\label{tab:hyper}
\scriptsize
\begingroup
\setlength{\tabcolsep}{2.5pt}\renewcommand{\arraystretch}{1.15}
\resizebox{\linewidth}{!}{%
\begin{tabular}{@{}lcccccccccccccccc>{\raggedright\arraybackslash}p{1.7cm}>{\raggedright\arraybackslash}p{1.9cm}@{}}
\toprule
Benchmark & Actor & Critic & $\gamma$ & $\lambda$ & Clip & KL & Ref. & Ent. & Epochs & Minib. & Envs & Rollout & Grad. & Warmup & Penalty & $\sigma_0$ & Actor input & Critic input\\
\midrule
SMAC, SMACv2 & 2e-5 & 3e-4 & 0.99 & 0.95 & 0.2 & 0.02 & 0 & 0 & 4 & 256 & 8 & 128 & 0.5 & 1st iter. & off & 0.1 & frozen recurrent features & state\\
MPE, MA-MuJoCo & 2e-5 & 3e-4 & 0.99 & 0.95 & 0.2 & 0.02 & 0 & 0 & 4 & 256 & 40 & 128 & 0.5 & 1st iter. & off & 0.1 & local observation & joint observation or state\\
\bottomrule
\end{tabular}}
\par\smallskip\raggedright
Actor, Critic: learning rates. $\gamma$: discount. $\lambda$: GAE parameter. Clip: PPO clipping. KL: trajectory KL threshold. Ref., Ent.: reference and entropy coefficients. Minib.: minibatch size. Envs: parallel environments. Rollout: rollout length. Grad.: gradient norm clipping. Warmup: critic warmup. Penalty: reference penalty. $\sigma_0$: initial Gaussian transition noise.
\endgroup
\end{table*}

%% file: tables/omiga_smacv2.tex
\begin{table*}[!hbp]
\centering
\caption{Published results and TeamFlowRL on OMIGA MuJoCo (continuous control) and SMACv2 (discrete control), with methods as rows and task and dataset pairs as columns. Notation follows Table~\ref{tab:main-results}. We collected the SMACv2 Good datasets ourselves and will release them. Published baselines report only Replay, so a TeamFlowRL result on Good is shaded and bold when it exceeds every published baseline on the same task. MAC-Flow$^{\mathrm{r}}$ is our local reproduction.}
\label{tab:omiga-smacv2}
\label{tab:mujoco_omiga}
\begingroup\scriptsize
\renewcommand{\tiny}{\fontsize{5}{6}\selectfont}
\setlength{\tabcolsep}{1.5pt}
\renewcommand{\arraystretch}{1.12}
\resizebox{\linewidth}{!}{%
\begin{tabular}{@{}lcccccccccc@{}}
\toprule
\multirow{2}{*}{\textbf{OMIGA}} & \multicolumn{2}{c}{\textbf{HalfCheetah}} & \multicolumn{3}{c}{\textbf{Hopper}} & \multicolumn{4}{c}{\textbf{Ant}} & \multirow{2}{*}{\textbf{Avg.}} \\
\cmidrule(lr){2-3}\cmidrule(lr){4-6}\cmidrule(lr){7-10}
 & Expert & M-E & Expert & M-R & M-E & Expert & Medium & M-R & M-E & \\
\midrule
MACQL~\citep{cql}~(\citeyear{cql}) & $4589.5$\,{\tiny$\pm 98.5$} & $4738.2$\,{\tiny$\pm 181.1$} & \cellcolor{baselinesecond}$\underline{3359.1}$\,{\tiny$\pm 513.8$} & $31.4$\,{\tiny$\pm 15.2$} & $2751.8$\,{\tiny$\pm 123.3$} & \cellcolor{baselinesecond}$\underline{2082.4}$\,{\tiny$\pm 21.7$} & $1033.9$\,{\tiny$\pm 66.4$} & $434.6$\,{\tiny$\pm 108.3$} & $1800.2$\,{\tiny$\pm 21.5$} & $2313.5$ \\
MATD3BC~\citep{td3bc}~(\citeyear{td3bc}) & $4401.6$\,{\tiny$\pm 169.1$} & $3518.1$\,{\tiny$\pm 381.0$} & $3309.9$\,{\tiny$\pm 4.5$} & $269.7$\,{\tiny$\pm 41.8$} & \cellcolor{baselinesecond}$\underline{2904.3}$\,{\tiny$\pm 477.4$} & $2046.9$\,{\tiny$\pm 17.1$} & $1422.6$\,{\tiny$\pm 21.1$} & $995.2$\,{\tiny$\pm 52.8$} & $1636.1$\,{\tiny$\pm 96.0$} & $2278.3$ \\
OMAR~\citep{omar}~(\citeyear{omar}) & $-206.7$\,{\tiny$\pm 161.1$} & $-253.8$\,{\tiny$\pm 63.9$} & $2.4$\,{\tiny$\pm 1.5$} & $3.3$\,{\tiny$\pm 3.2$} & $1.4$\,{\tiny$\pm 0.9$} & $312.5$\,{\tiny$\pm 297.5$} & $-1710.0$\,{\tiny$\pm 1589.0$} & $-2014.2$\,{\tiny$\pm 844.7$} & $-2992.8$\,{\tiny$\pm 7.0$} & $-762.0$ \\
OMIGA~\citep{omiga}~(\citeyear{omiga}) & $3383.6$\,{\tiny$\pm 552.7$} & $2948.5$\,{\tiny$\pm 518.9$} & $859.6$\,{\tiny$\pm 709.5$} & $774.2$\,{\tiny$\pm 494.3$} & $709.0$\,{\tiny$\pm 595.7$} & $2055.5$\,{\tiny$\pm 1.6$} & $1418.4$\,{\tiny$\pm 5.4$} & $1105.1$\,{\tiny$\pm 88.9$} & $1720.3$\,{\tiny$\pm 110.6$} & $1663.8$ \\
MADiff-D~\citep{madiff}~(\citeyear{madiff}) & \cellcolor{baselinesecond}$\underline{4711.4}$\,{\tiny$\pm 213.6$} & $4410.9$\,{\tiny$\pm 836.8$} & $2853.3$\,{\tiny$\pm 593.8$} & $936.1$\,{\tiny$\pm 574.0$} & $2810.4$\,{\tiny$\pm 723.2$} & $2060.0$\,{\tiny$\pm 10.3$} & $1428.4$\,{\tiny$\pm 14.7$} & $1294.5$\,{\tiny$\pm 360.2$} & $1740.2$\,{\tiny$\pm 158.9$} & $2471.69$ \\
MAC-Flow~\citep{macflow}~(\citeyear{macflow}) & $4650.0$\,{\tiny$\pm 271.6$} & \cellcolor{baselinesecond}$\underline{5139.9}$\,{\tiny$\pm 84.1$} & \cellcolor{baselinebest}$\mathbf{3592.1}$\,{\tiny$\pm 8.9$} & \cellcolor{baselinesecond}$\underline{1166.3}$\,{\tiny$\pm 451.9$} & \cellcolor{baselinebest}$\mathbf{2988.3}$\,{\tiny$\pm 480.2$} & $2060.2$\,{\tiny$\pm 20.0$} & \cellcolor{baselinesecond}$\underline{1432.4}$\,{\tiny$\pm 17.8$} & \cellcolor{baselinesecond}$\underline{1498.4}$\,{\tiny$\pm 20.3$} & \cellcolor{baselinesecond}$\underline{2053.3}$\,{\tiny$\pm 20.4$} & \cellcolor{baselinesecond}$\underline{2731.2}$ \\
\midrule
TeamFlowRL (w/o online RL) & $3477.40$\,{\tiny$\pm 246.94$} & $3021.08$\,{\tiny$\pm 128.55$} & $2896.91$\,{\tiny$\pm 48.37$} & $403.84$\,{\tiny$\pm 37.51$} & $900.09$\,{\tiny$\pm 64.63$} & $2048.93$\,{\tiny$\pm 6.48$} & $1420.26$\,{\tiny$\pm 3.02$} & $1092.11$\,{\tiny$\pm 19.17$} & $1768.56$\,{\tiny$\pm 63.84$} & $1892.13$ \\
\textbf{TeamFlowRL} & \cellcolor{baselinebest}$\mathbf{6562.96}^{\ddagger}$\,{\tiny$\pm 346.62$} & \cellcolor{baselinebest}$\mathbf{6828.97}$\,{\tiny$\pm 91.78$} & $3059.66^{\ddagger}$\,{\tiny$\pm 116.38$} & \cellcolor{baselinebest}$\mathbf{3287.96}$\,{\tiny$\pm 138.76$} & $2055.42$\,{\tiny$\pm 102.72$} & \cellcolor{baselinebest}$\mathbf{3867.80}$\,{\tiny$\pm 27.18$} & \cellcolor{baselinebest}$\mathbf{3874.97}$\,{\tiny$\pm 35.42$} & \cellcolor{baselinebest}$\mathbf{3761.87}$\,{\tiny$\pm 23.21$} & \cellcolor{baselinebest}$\mathbf{5193.92}$\,{\tiny$\pm 108.55$} & \cellcolor{baselinebest}$\mathbf{4277.06}$ \\
\bottomrule
\end{tabular}}
\par\vspace{3pt}
\begin{tabular*}{\linewidth}{@{\extracolsep{\fill}}lcccccc@{}}
\toprule
\multirow{2}{*}{\textbf{SMACv2}} & \multicolumn{2}{c}{\textbf{Terran 5v5}} & \multicolumn{2}{c}{\textbf{Zerg 5v5}} & \multicolumn{1}{c}{\textbf{Terran 10v10}} & \multirow{2}{*}{\textbf{Avg.}} \\
\cmidrule(lr){2-3}\cmidrule(lr){4-5}\cmidrule(lr){6-6}
 & Replay & Good & Replay & Good & Replay & \\
\midrule
BC~\citep{bc}~(\citeyear{bc}) & $7.3$\,{\tiny$\pm 1.0$} & -- & $6.8$\,{\tiny$\pm 0.6$} & -- & $7.4$\,{\tiny$\pm 0.5$} & $7.2$ \\
MABCQ~\citep{bcq}~(\citeyear{bcq}) & \cellcolor{baselinesecond}$\underline{13.8}$\,{\tiny$\pm 4.4$} & -- & \cellcolor{baselinesecond}$\underline{10.3}$\,{\tiny$\pm 1.2$} & -- & $12.7$\,{\tiny$\pm 2.0$} & $12.3$  \\
MACQL~\citep{cql}~(\citeyear{cql}) & $11.8$\,{\tiny$\pm 0.9$} & -- & \cellcolor{baselinesecond}$\underline{10.3}$\,{\tiny$\pm 3.4$} & -- & $11.8$\,{\tiny$\pm 2.0$} & $11.3$  \\
Diffusion BC~\citep{diffusion_policy}~(\citeyear{diffusion_policy}) & $9.3$\,{\tiny$\pm 0.9$} & -- & $8.1$\,{\tiny$\pm 1.7$} & -- & $5.5$\,{\tiny$\pm 1.5$} & $7.6$ \\
Flow BC~\citep{flow}~(\citeyear{flow}) & $8.3$\,{\tiny$\pm 1.9$} & -- & $4.6$\,{\tiny$\pm 0.5$} & -- & $5.8$\,{\tiny$\pm 1.7$} & $6.2$ \\
MADiff-D~\citep{madiff}~(\citeyear{madiff}) & $13.3$\,{\tiny$\pm 1.8$} & -- & $10.2$\,{\tiny$\pm 1.1$} & -- & \cellcolor{baselinebest}$\mathbf{13.8}$\,{\tiny$\pm 1.3$} & \cellcolor{baselinesecond}$\underline{12.4}$  \\
MAC-Flow~\citep{macflow}~(\citeyear{macflow}) & \cellcolor{baselinebest}$\mathbf{16.6}$\,{\tiny$\pm 4.3$} & -- & $9.8$\,{\tiny$\pm 1.5$} & -- & \cellcolor{baselinesecond}$\underline{13.0}$\,{\tiny$\pm 4.7$} & \cellcolor{baselinebest}$\mathbf{13.1}$  \\
MAC-Flow$^{\mathrm{r}}$ & $9.70$\,{\tiny$\pm 1.40$} & -- & $7.01$\,{\tiny$\pm 1.27$} & -- & $6.03$\,{\tiny$\pm 0.57$} & $7.58$ \\
\midrule
TeamFlowRL (w/o online RL) & $7.22$\,{\tiny$\pm 0.74$} & $14.22$\,{\tiny$\pm 0.45$} & $6.24$\,{\tiny$\pm 0.56$} & $12.42$\,{\tiny$\pm 0.37$} & $4.77$\,{\tiny$\pm 0.94$} & $6.08$ \\
\textbf{TeamFlowRL} & $12.86$\,{\tiny$\pm 0.78$} & $15.17$\,{\tiny$\pm 0.81$} & \cellcolor{baselinebest}$\mathbf{10.86}$\,{\tiny$\pm 1.18$} & \cellcolor{baselinebest}$\mathbf{14.17}$\,{\tiny$\pm 1.00$} & $10.16$\,{\tiny$\pm 0.96$} & $11.29$\,{\tiny$\pm 0.93$}  \\
\bottomrule
\end{tabular*}
\endgroup
\par\smallskip
{\footnotesize\raggedright $\ddagger$ bounded repair: the transition mean is bounded as $10\cdot\tanh(\mathrm{mean}/10)$, a structural change for numerical stability.  SMACv2 averages are over the three Replay tasks.
\par}
\end{table*}

%% file: tables/robot_configs.tex
\begin{table*}[!hbp]
\centering\scriptsize
\caption{Configurations for the robot experiments with state observations.}
\label{tab:robot-configs}
\setlength{\tabcolsep}{2pt}\renewcommand{\arraystretch}{1.1}
\resizebox{\linewidth}{!}{%
\begin{tabular}{@{}llccccccccccccccccccc@{}}
\toprule
 & & \multicolumn{6}{c}{Pretraining} & \multicolumn{13}{c}{Online fine-tuning}\\
\cmidrule(lr){3-8}\cmidrule(lr){9-21}
& Method & Init. & Pre.\ upd. & Demos & Split & Batch & Seq. & Steps & Envs & Ep.\ len. & Actor & Critic & Epochs & Clip & KL stop & Ref.\ KL & Ent. & Minib. & Rollout & Exec.\\
\midrule
\multirow{2}{*}{MS3} & TeamFlowRL & 10-step & 50k & 983/1007 & 90/10 & 128 & 20 & 5M & 64 & 100 & 2e-5 & 3e-4 & 4 & -- & 0.02 & -- & -- & -- & 6400 & random $z$\\
 & BC+MAPPO & 1-step & 50k & 983/1007 & 90/10 & 128 & 20 & 5M & 64 & 100 & 2e-5 & 3e-4 & 4 & 0.05 & 0.02 & 0.01 & 0.001 & 128 & 6400 & mean\\
\midrule
\multirow{2}{*}{RF} & TeamFlowRL & 10-step & 30k & 817/1024 & 735/82 & -- & -- & 2M & 64 & 100 & 2e-5 & 3e-4 & 4 & -- & 0.02 & -- & -- & -- & 6400 & --\\
 & BC+MAPPO & 1-step & 30k & 817/1024 & 735/82 & -- & -- & 2M & 64 & 100 & 2e-5 & 3e-4 & 4 & -- & 0.02 & -- & -- & -- & 6400 & mean\\
\bottomrule
\end{tabular}}
\par\smallskip\raggedright MS3: ManiSkill3. RF: RoboFactory. Init.: initial policy. Actor, Critic: learning rates. Pre.\ upd.: pretraining updates. Split: train/validation. Seq.: pretraining sequence length. Steps: online environment steps. Ep.\ len.: episode length. Ref.\ KL: reference KL coefficient. Ent.: entropy coefficient. Minib.: online minibatch. Exec.: execution rule. The two methods share the pretraining run and demonstration split within each benchmark. ManiSkill3 demonstrations are listed for Pick / Stack, and RoboFactory uses 817 successful of 1,024 expert episodes  and the pretraining checkpoint at 30k of 50k updates. The TeamFlowRL execution rule samples a random initial latent without transition noise.  Rollout batch sizes count joint environment transitions.
\end{table*}

%% file: release_materials/release_section.tex
\subsection{Learning to Release and Withdraw}
\label{sec:release_withdraw}
The hands-away variant of the two-arm stacking tasks in ManiSkill and RoboFactory requires the arms to release the blocks, withdraw, and leave the completed stack standing. An episode succeeds when the original task succeeds and, at the end of the episode, neither arm touches the blocks and both hands are away from them.

In ManiSkill, training starts from the pretrained stacking policies and adds a reward term for withdrawal. In RoboFactory, the original demonstrations contain no withdrawal, so the flow policies are pretrained on new demonstrations that include it. The thresholds and remaining settings are provided in the released code.